\documentclass[]{worldbench}
\definecolor{w_blue}{RGB}{52,204,204}
\definecolor{w_yellow}{RGB}{255,192,0}

\usepackage{amsfonts}
\usepackage{amsmath}
\usepackage{amssymb}

\usepackage{colortbl}

\usepackage{makecell}
\usepackage{multicol}
\usepackage{multirow}
\usepackage{pifont}

\definecolor{red}{rgb}{0.8,0,0}  
\definecolor{green}{RGB}{0, 133, 21}  
\definecolor{grey}{rgb}{0.5,0.5,0.5}

\definecolor{w_1}{RGB}{52,204,204}
\definecolor{w_2}{RGB}{70,203,187}
\definecolor{w_3}{RGB}{95,202,161}
\definecolor{w_4}{RGB}{119,200,136}
\definecolor{w_5}{RGB}{155,199,101}
\definecolor{w_6}{RGB}{185,197,70}
\definecolor{w_7}{RGB}{227,194,28}
\definecolor{w_8}{RGB}{243,193,12}
\definecolor{w_9}{RGB}{255,192,0}

\newcommand{\ours}{\textbf{\textsf{\textcolor{3eed_green}{3}\textcolor{3eed_red}{E}\textcolor{3eed_blue}{E}\textcolor{gray}{D}}}}

\makeatletter
\def\blfootnote{\xdef\@thefnmark{}\@footnotetext}
\DeclareRobustCommand\onedot{\futurelet\@let@token\@onedot}
\def\@onedot{\ifx\@let@token.\else.\null\fi\xspace}
\def\eg{\textit{e.g}\onedot}

\def\ie{\textit{i.e}\onedot}

\def\vs{\textit{vs}\onedot}

\makeatother

\def\eqref#1{Equation~\ref{#1}}

\definecolor{3eed_red}{RGB}{239,99,75}
\definecolor{3eed_blue}{RGB}{99,113,250}
\definecolor{3eed_green}{RGB}{0,180,139}
\definecolor{3eed_yellow}{RGB}{229,157,35}
\definecolor{3eed_gray}{RGB}{165,165,165}
\definecolor{3eedlink}{RGB}{0,180,139}

\def\ie{\textit{i.e.}}

\def\vs{\textit{vs.}}

\def\eg{\textit{e.g.}}

\usepackage{pifont}
\newcommand{\cmark}{\ding{51}}
\newcommand{\xmark}{\ding{55}}

\usepackage{soul}

\usepackage{titletoc}
\usepackage{bbm}

\title{\textsc{3EED}: Ground Everything Everywhere in 3D}

\author[]{Rong Li~\raisebox{0.2em}{\includegraphics[width=0.019\linewidth]{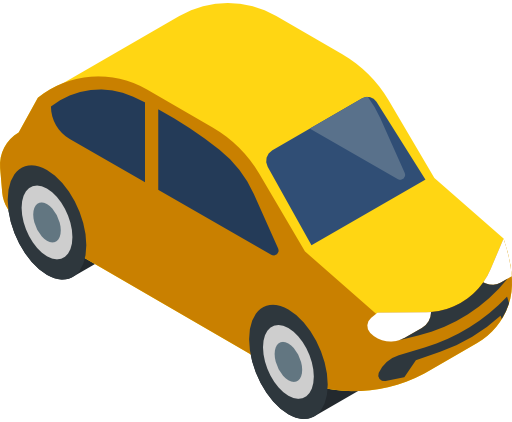}}}
\author[]{Yuhao Dong~\raisebox{0.2em}{\includegraphics[width=0.019\linewidth]{figures/icons/car1.png}}}
\author[]{Tianshuai Hu~\raisebox{0.2em}{\includegraphics[width=0.019\linewidth]{figures/icons/car1.png}}}
\author[]{Ao Liang~\raisebox{0.2em}{\includegraphics[width=0.019\linewidth]{figures/icons/car1.png}}}
\author[]{Youquan Liu~\raisebox{0.2em}{\includegraphics[width=0.019\linewidth]{figures/icons/car1.png}}}
\author[]{Dongyue Lu~\raisebox{0.2em}{\includegraphics[width=0.019\linewidth]{figures/icons/car1.png}}}
\author[]{Liang Pan}
\author[]{Lingdong Kong~\raisebox{0.2em}{\includegraphics[width=0.019\linewidth]{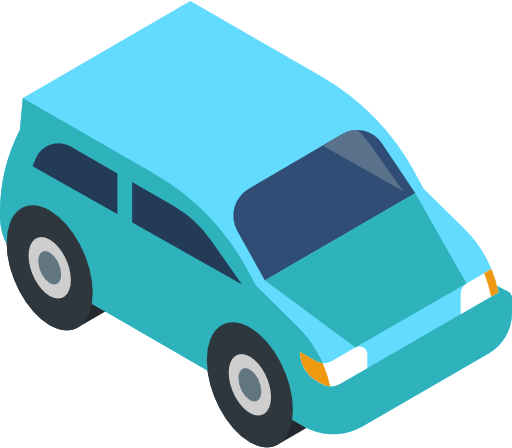}}}
\author[]{Junwei Liang~\raisebox{0.15em}{\includegraphics[width=0.017\linewidth]{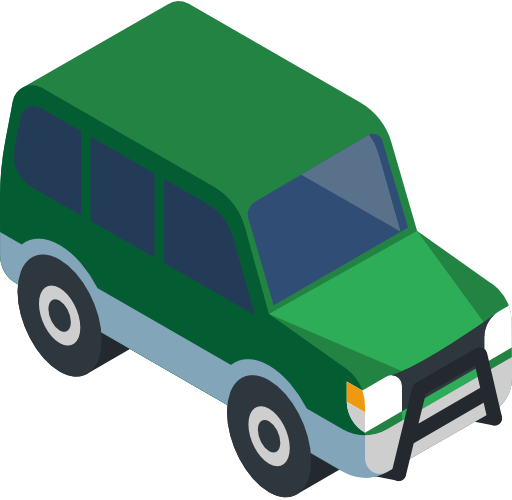}}}
\author[]{Ziwei Liu~\raisebox{0.15em}{\includegraphics[width=0.017\linewidth]{figures/icons/car4.png}}}

\affiliation[]{
\raisebox{-0.1em}{\includegraphics[width=0.029\linewidth]{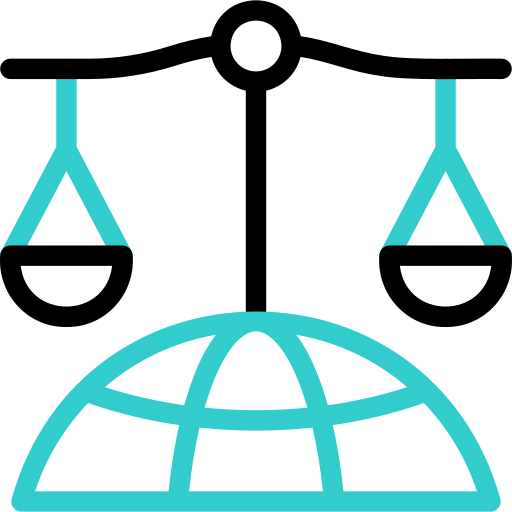}}~WorldBench Team
\\[1.2ex]
~\raisebox{-0.2em}{\includegraphics[width=0.032\linewidth]{figures/icons/car1.png}}~{\small \textbf{Equal Contributions}}
\quad
\raisebox{-0.2em}{\includegraphics[width=0.031\linewidth]{figures/icons/car2.png}}~{\small \textbf{Project Lead}}
\quad
\raisebox{-0.2em}{\includegraphics[width=0.028\linewidth]{figures/icons/car4.png}}~{\small \textbf{Corresponding Author}}
}

\abstract{
Visual grounding in 3D is the key for embodied agents to localize language-referred objects in open-world environments. However, existing benchmarks are limited to indoor focus, single-platform constraints, and small scale. We introduce \ours, a \emph{multi-platform}, \emph{multi-modal} 3D grounding benchmark featuring RGB and LiDAR data from \textbf{vehicle}, \textbf{drone}, and \textbf{quadruped} platforms. We provide over $128{,}000$ objects and $22{,}000$ validated referring expressions across diverse outdoor scenes -- $\mathbf{10\times}$ larger than existing datasets. We develop a scalable annotation pipeline combining vision-language model prompting with human verification to ensure high-quality spatial grounding. To support cross-platform learning, we propose platform-aware normalization and cross-modal alignment techniques, and establish benchmark protocols for \textbf{in-domain} and \textbf{cross-platform} 3D grounding evaluations. Our findings reveal significant performance gaps, highlighting the challenges and opportunities of generalizable 3D grounding. The \ours~dataset and benchmark toolkit have been released to advance future research in language-driven 3D embodied perception.
}

\metadata[
\raisebox{-0.2em}{\includegraphics[width=0.025\linewidth]{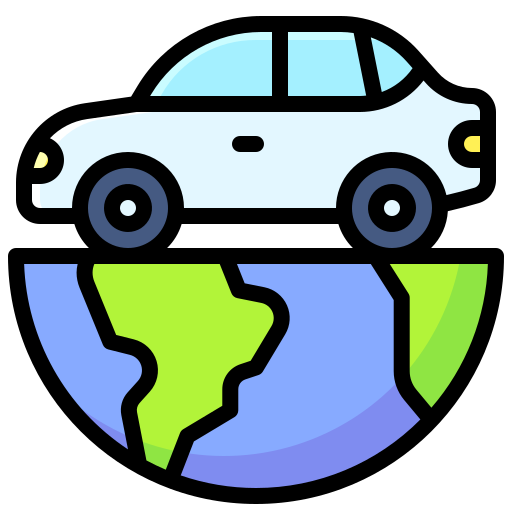}}~~Project Page]{\href{https://project-3eed.github.io/}{\texttt{https://project-3eed.github.io}}
\\[-1.5ex]}

\metadata[
\raisebox{-0.2em}{\includegraphics[width=0.025\linewidth]{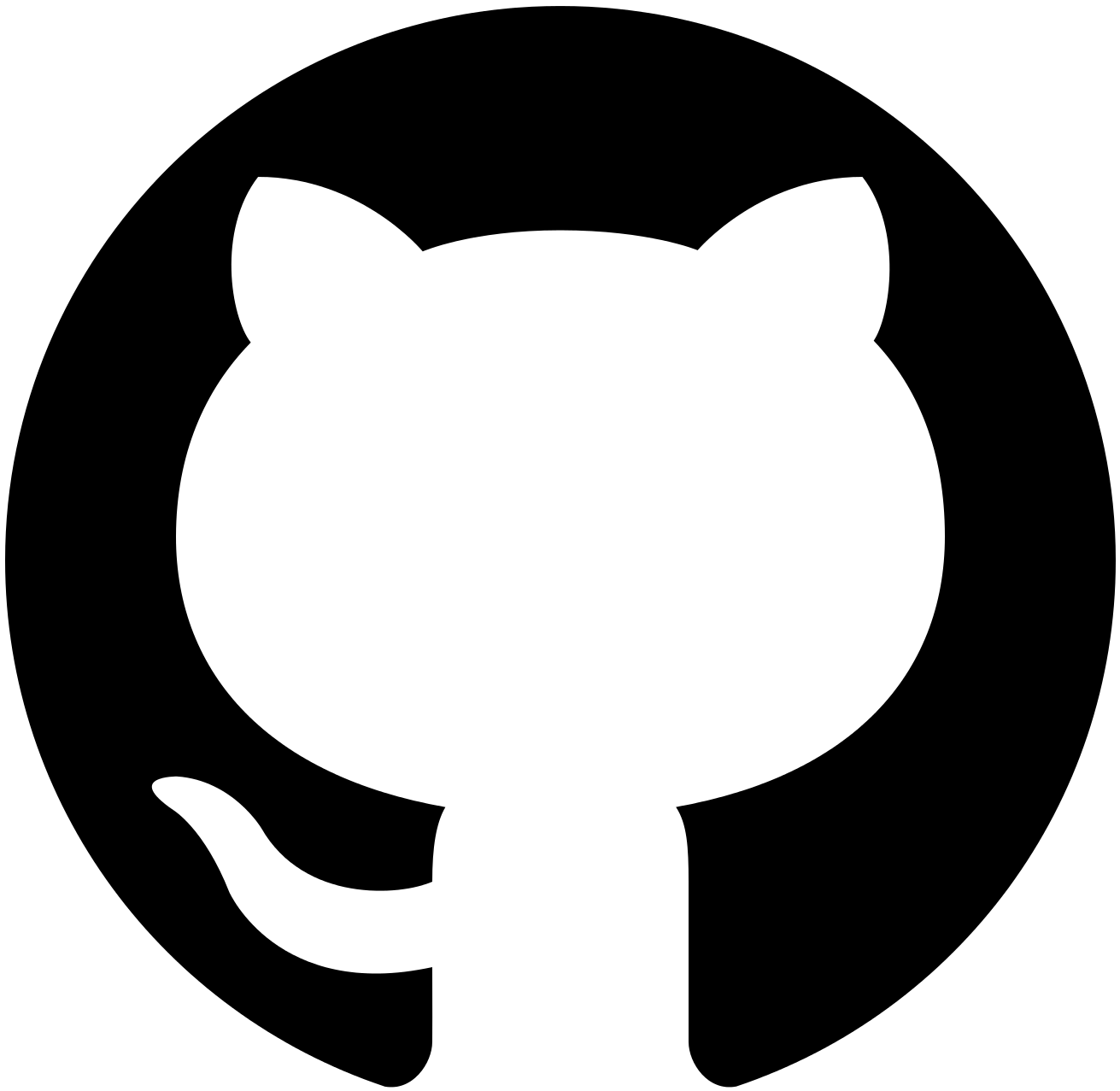}}~~GitHub Repo]{\href{https://github.com/worldbench/3EED}{\texttt{https://github.com/worldbench/3EED}}
\\[-1.7ex]}

\metadata[
\raisebox{-0.3em}{\includegraphics[width=0.026\linewidth]{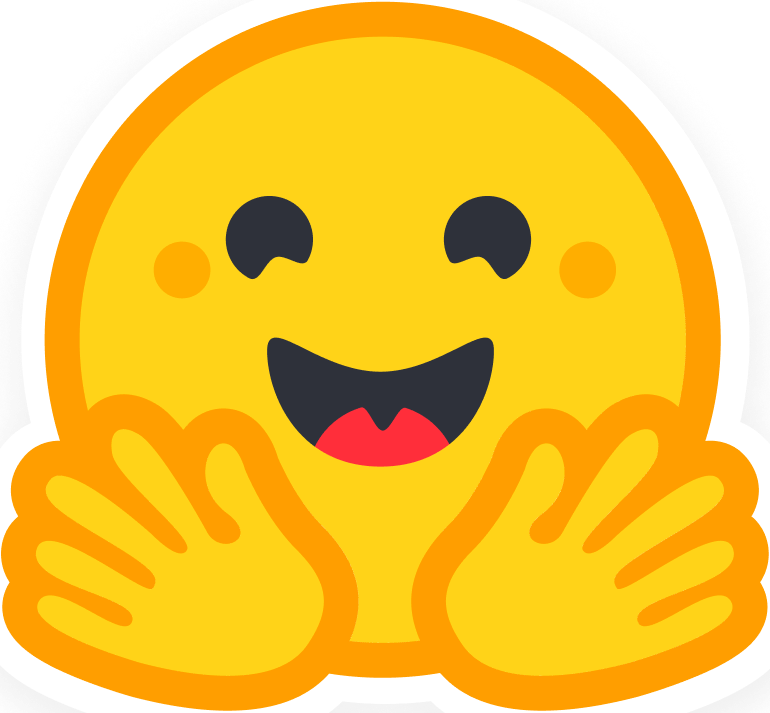}}~~HuggingFace Dataset]{\href{https://huggingface.co/datasets/3EED/3EED}{\texttt{https://huggingface.co/datasets/3EED}}}

\begin{document}

\maketitle

\begin{figure}[h]
    \centering
    \vspace{0.1cm}
    \includegraphics[width=\linewidth]{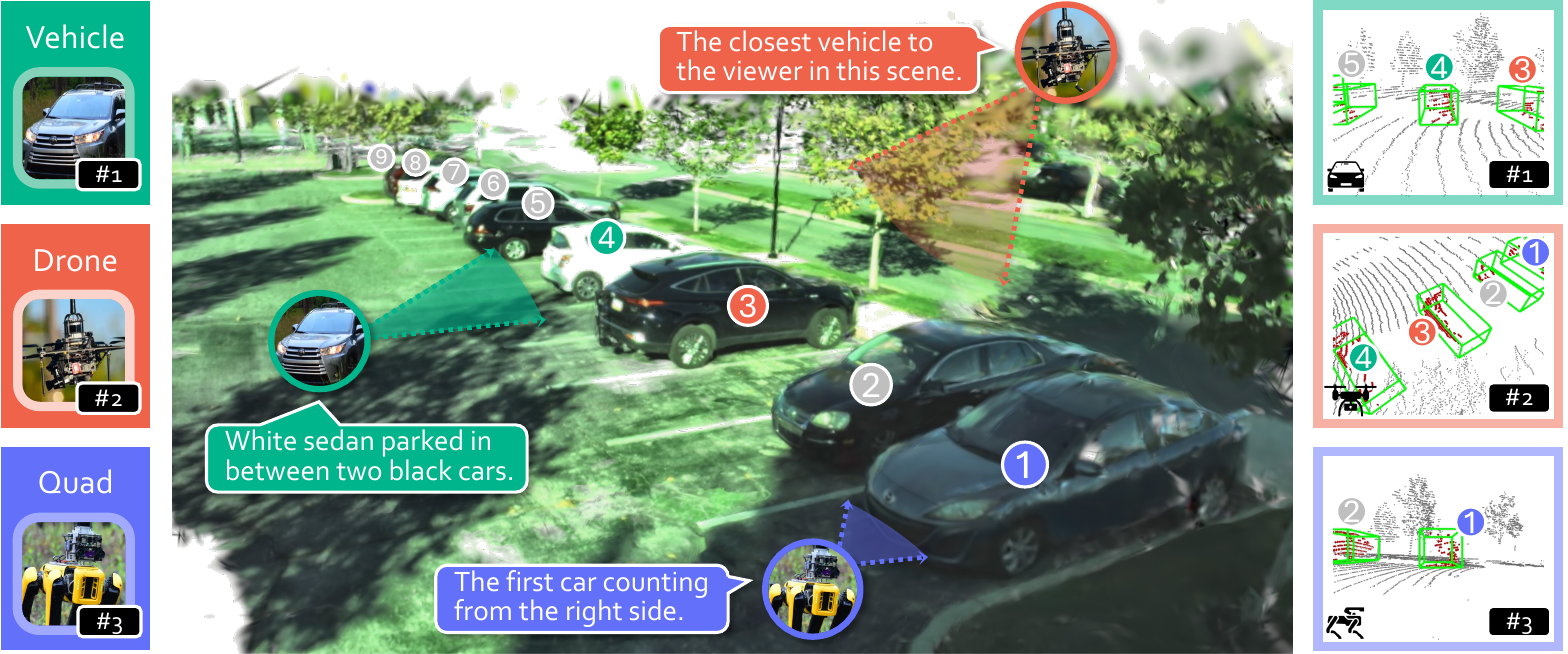}
    \vspace{-0.6cm}
    \caption{\textbf{Multi-modal, multi-platform 3D grounding from} \ours. Given a scene and a structured natural language expression, the task is to localize the referred object in 3D space. Our dataset captures diverse embodied robot sensing viewpoints from {\includegraphics[width=0.021\linewidth]{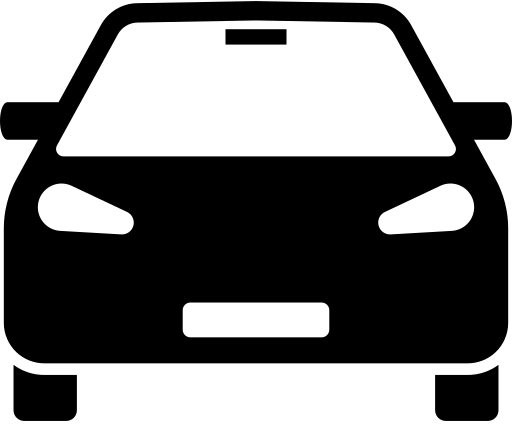}} \texttt{Vehicle}, {\includegraphics[width=0.028\linewidth]{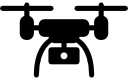}} \texttt{Drone}, and {\includegraphics[width=0.022\linewidth]{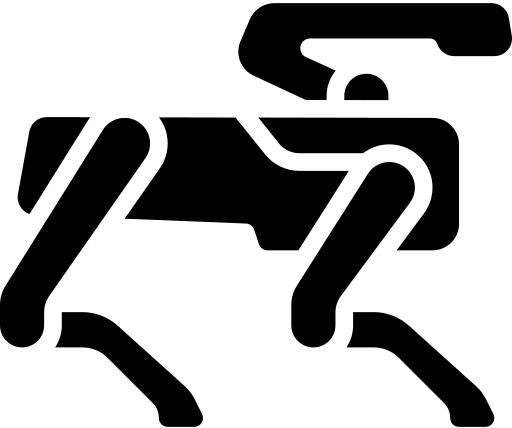}} \texttt{Quadruped} platforms, presenting unique challenges in spatial reasoning, scene analysis, and cross-platform 3D generalization.}
    \label{fig:teaser}
    \vspace{-0.55cm}
\end{figure}

\section{Introduction}
\label{sec:intro}

Grounding free-form language to 3D scenes is a core capability for embodied agents operating in the physical world~\cite{achlioptas2020referit3d, chen2020scanrefer, behley2019semanticKITTI, boulch2023also, survey_3d_4d_world_models}. By associating natural language expressions with physical objects in 3D space, robots and autonomous systems can interpret high-level human instructions to perform downstream tasks, \eg, navigation, interaction, and situational awareness~\cite{pendleton2017perception, xiao2023survey, zeng2024self, fong2022panoptic-nuScenes, pan2020semanticPOSS, xiao2023semanticSTF, xiao2022synLiDAR}.

Recent advances in 3D visual grounding have primarily focused on indoor benchmarks \cite{jain2022butd-detr, bakr2022lar, huang2022mvt}, where sensing is constrained, scenes are small, and objects are limited to household categories \cite{yang2024llmgrounder, yuan2024zsvg3d}. However, real-world applications require models to operate in outdoor environments with greater spatial scale \cite{milioto2019rangenet++, kong2023rethinking, liu2025lalalidar}, diverse viewpoints \cite{puy23waffleiron, cheng2022cenet, liang2025pi3det}, and sparse sensor data~\cite{baur2025liso, kong2023robo3d, liang2025lidarcrafter}.

While recent datasets have begun addressing outdoor 3D grounding~\cite{jiang2021rellis3D, gao2021survey, xiao2023survey, hao2024is}, they remain limited by single-platform data (\eg, vehicle-mounted LiDAR), small scale with few objects and expressions, and a lack of multi-modal supervision, often providing only LiDAR or RGB but not both~\cite{hong20224dDSNet, li2024rapid, hu2020randla, liong2020amvnet, jaritz2020xMUDA, liu2023uniseg, li2024is, wang2025monomrn, xu2025lima}. These gaps limit the development of models that generalize across platforms, modalities, and real-world conditions.

To address these gaps, we introduce \ours, a \emph{large-scale, multi-platform, multi-modal} benchmark for 3D visual grounding in outdoor environments (see Figure~\ref{fig:teaser}). Our dataset captures synchronized LiDAR and RGB data from three distinct robotic platforms: {\includegraphics[width=0.021\linewidth]{figures/icons/vehicle.png}} \texttt{Vehicle}, {\includegraphics[width=0.028\linewidth]{figures/icons/drone.png}} \texttt{Drone}, {\includegraphics[width=0.022\linewidth]{figures/icons/quadruped.png}} \texttt{Quadruped}. It provides over $\mathbf{128{,}000}$ \textbf{object instances} and $\mathbf{22{,}000}$ human-verified \textbf{referring expressions}, making it $\mathbf{10\times}$ \textbf{larger} than existing outdoor grounding benchmarks, as compared in Table~\ref{tab:dataset_comparison}.

To enable scalable annotation, we develop a \emph{vision-language model prompting pipeline} combined with \emph{human-in-the-loop verification} to generate high-quality referring expressions. Additionally, we propose \textbf{platform-aware normalization} and \textbf{cross-modal alignment} techniques to standardize geometric and sensory data while preserving platform-specific characteristics. Based on these contributions, we establish a comprehensive benchmark suite covering in-domain, cross-platform, and multi-object grounding settings. Through extensive experiments with state-of-the-art models \cite{jain2022butd-detr,wu2023eda}, we reveal substantial performance gaps across platforms, exposing the challenges of robust and generalizable 3D visual grounding in real-world outdoor environments.

To summarize, the key contributions of this work to the related fields include:
\begin{itemize}
    \item We present \ours, the first large-scale, multi-platform, multi-modal 3D visual grounding benchmark spanning {\includegraphics[width=0.021\linewidth]{figures/icons/vehicle.png}} \texttt{Vehicle}, {\includegraphics[width=0.028\linewidth]{figures/icons/drone.png}} \texttt{Drone}, {\includegraphics[width=0.022\linewidth]{figures/icons/quadruped.png}} \texttt{Quadruped} platforms, covering over $128{,}000$ objects and $22{,}000$ human-verified expressions, which is $10\times$ larger than existing outdoor datasets.
    
    \item We develop a scalable annotation pipeline combining vision-language model prompting with human validation, enabling high-quality and diverse language supervision.
    
    \item We propose \emph{platform-aware normalization} and \emph{cross-modal alignment} to unify sensor geometry and synchronize LiDAR, RGB, and language cues, enabling consistency across diverse platforms.
    
    \item We establish comprehensive benchmark protocols for in-domain, cross-platform, and multi-object grounding, along with strong baseline evaluations revealing key challenges and future directions.
\end{itemize}

\begin{table}[t]
    \centering
    \caption{\textbf{Summary of existing outdoor 3D visual grounding benchmarks.} We compare key features from aspects including: $^1$\textbf{Platform} ({\includegraphics[width=0.021\linewidth]{figures/icons/vehicle.png}} \texttt{Vehicle}, {\includegraphics[width=0.028\linewidth]{figures/icons/drone.png}} \texttt{Drone}, and {\includegraphics[width=0.022\linewidth]{figures/icons/quadruped.png}} \texttt{Quadruped}), $^2$\textbf{Area Coverage}, and $^3$\textbf{Statistics}. Our dataset exhibits advantages in platform diversity, large collections of LiDAR (\textbf{L}) and camera (\textbf{C}) scenes (\textbf{Sce.}), 3D objects (\textbf{Obj.}), referring expressions (\textbf{Expr.}), and rich elevation variations (\textbf{Elev.}).
    }
    \vspace{-0.2cm}
    \resizebox{\linewidth}{!}{
    \begin{tabular}{r|c|ccc|c|cccc}
    \toprule
    \multirow{2}{*}{\textbf{Dataset}} & \multirow{2}{*}{\textbf{Sensor}} & \multicolumn{3}{c|}{\textbf{Platform}} & \textbf{Scene} & \multicolumn{4}{c}{\textbf{Statistics}}
    \\
    & & \raisebox{-0.2\height}{\includegraphics[width=0.026\linewidth]{figures/icons/vehicle.png}} & \raisebox{-0.2\height}{\includegraphics[width=0.037\linewidth]{figures/icons/drone.png}}  & \raisebox{-0.2\height}{\includegraphics[width=0.027\linewidth]{figures/icons/quadruped.png}} & \textbf{Coverage} & \textbf{\#Sce.} & \textbf{\#Obj.} & \textbf{\#Expr.} & \textbf{\#Elev.}
    \\
    \midrule\midrule

    Mono3DRefer~\cite{zhan2024mono3drefer} & C & \textcolor{3eed_green}{\cmark} & \textcolor{3eed_red}{\xmark} & \textcolor{3eed_red}{\xmark} & $140\text{m} \times 140\text{m}$  & $2{,}025$ & $8{,}228$ & $41{,}140$ & $42.8\text{m}$
    \\
    KITTI360Pose \cite{kolmet2022text2pos_kitti360pose} & L & \textcolor{3eed_green}{\cmark} & \textcolor{3eed_red}{\xmark} & \textcolor{3eed_red}{\xmark} & $140\text{m} \times 140\text{m}$  &  -  & $14{,}934$ & $43{,}381$ & $42.8\text{m}$
    \\
    CityRefer \cite{miyanishi2023cityrefer}  & L & \textcolor{3eed_red}{\xmark} & \textcolor{3eed_green}{\cmark} & \textcolor{3eed_red}{\xmark} & - & - & $5{,}866$ & $35{,}196$ & -
    \\
    STRefer \cite{lin2024wildrefer} & L + C & \textcolor{3eed_green}{\cmark} & \textcolor{3eed_red}{\xmark} & \textcolor{3eed_red}{\xmark} & $60\text{m} \times 60\text{m}$  & $662$ & $3{,}581$ & $5{,}458$ & -
    \\
    LifeRefer \cite{lin2024wildrefer} & L + C & \textcolor{3eed_green}{\cmark} & \textcolor{3eed_red}{\xmark} & \textcolor{3eed_red}{\xmark} & $60\text{m} \times 60\text{m}$  & $3{,}172$ & $11{,}864$ & $25{,}380$ & -
    \\
    Talk2LiDAR \cite{liu2024talk2lidar} & L + C & \textcolor{3eed_green}{\cmark} & \textcolor{3eed_red}{\xmark} & \textcolor{3eed_red}{\xmark} & $140\text{m} \times 140\text{m}$  & $6{,}419$ & - & $59{,}207$ & $48.6\text{m}$
    \\
    Talk2Car-3D \cite{baek2024lidarefer_talk2car3d}  & L + C & \textcolor{3eed_green}{\cmark} & \textcolor{3eed_red}{\xmark} & \textcolor{3eed_red}{\xmark} & $140\text{m} \times 140\text{m}$ & $5{,}534$ & - & $10{,}169$ & $48.6\text{m}$
    \\
    \rowcolor{3eed_green!10}
    \ours~\textbf{\small(Ours)} & L + C & \textcolor{3eed_green}{\cmark} & \textcolor{3eed_green}{\cmark} & \textcolor{3eed_green}{\cmark} & $\mathbf{280\text{m} \times 240\text{m}}$ & $\mathbf{20{,}367}$  & $\mathbf{128{,}735}$ & $\mathbf{22{,}439}$ &  $\mathbf{80}$\text{m}
    \\
    \bottomrule
\end{tabular}
}
\label{tab:dataset_comparison}
\end{table}

\section{Related Work}
\label{sec:related_work}

\noindent\textbf{3D Visual Grounding.}
3D visual grounding localizes objects in 3D scenes from natural language expressions. Early efforts focus on indoor RGB-D datasets like ScanRefer \cite{chen2020scanrefer} and Nr3D \cite{achlioptas2020referit3d}, built on ScanNet \cite{dai2017scannet} and ARKitScenes \cite{baruch2021arkitscenes}, with object categories mostly limited to furniture. Recent datasets such as Multi3DRefer \cite{zhang2023multi3drefer} and EmbodiedScan \cite{wang2024embodiedscan} expand to multi-object and egocentric grounding. These resources have driven the development of various models \cite{zhao20213dvg, yang2021sat, wu2023eda, guo2023viewrefer, unal2024concretenet, jain2022butd-detr, bakr2022lar, huang2022mvt, wang2024g3lq, yuan2024zsvg3d, zhu20233dvista-scanscribe, li2024seeground, yang2024llmgrounder} focused on spatial-linguistic alignment in controlled indoor environments.

\noindent\textbf{3D Grounding in the Wild.}
Grounding language in outdoor 3D scenes introduces challenges such as large spatial scales, sparse point clouds, and diverse object distributions~\cite{kong2023lasermix,kong2025lasermix2,xiao2022polarmix,tang2022torchsparse,tang2023torchsparse++,razani2021lite}. Talk2Car~\cite{deruyttere2019talk2car}, based on nuScenes~\cite{caesar2020nuscenes}, is an early benchmark for driving scenarios. STRefer~\cite{lin2024wildrefer} extends this with RGB and LiDAR from mobile agents, focusing on human activities. Mono3DVG~\cite{zhan2024mono3drefer} studies grounding in monocular images without 3D sensors. KITTI360Pose~\cite{kolmet2022text2pos_kitti360pose} uses templated language for text-to-position grounding in KITTI-360~\cite{geiger2012kitti}, targeting positions rather than objects. Talk2LiDAR~\cite{liu2024talk2lidar} and CityRefer~\cite{miyanishi2023cityrefer} provide multi-sensor and city-scale grounding tasks. However, all these datasets are limited to \textbf{single-platform} data acquisition.

\noindent\textbf{Language-Guided Perception in Embodied Platforms.}
Language understanding has also been explored in interactive~\cite{wang2024interact, linghu2024situated-resason, majumdar2022zson, sg3d, hong2021vln,dong2024insight,liu2024chain} and multi-task perception settings~\cite{zhu20233dvista-scanscribe, chen2023unit3d, huang2023embodied_generalist, yang2024octopus,yang2025egolife,jia2024sceneverse,xu2025frnet,liu2024multi,liu2021ppkt,sautier2022slidr,mahmoud2023st,liu2023seal,xu2024superflow}. Refer-KITTI~\cite{wu2023referkitti} based on KITTI~\cite{geiger2012kitti} enables tracking multiple objects with a single prompt. nuPrompt~\cite{wu2025nu-prompt} employs a language prompt to predict the object trajectory across views and frames. nuScenes-QA \cite{qian2024nuscenes-qa} formulates a multi-modal question answering benchmark using nuScenes \cite{caesar2020nuscenes} data. DriveLM \cite{sima2024drivelm} formulates driving as a graph-based visual question answering task, leveraging structured visual representations and large language models~\cite{gpt35} to answer route-planning and scene-understanding queries. These methods, however, focus on vehicle-based data~\cite{geiger2012kitti, caesar2020nuscenes} and semantic-level tasks~\cite{schult2022mask3d, huang2021text}, whereas our dataset enables fine-grained 3D grounding across diverse embodied agents, including drones and legged robots.

\section{\ours: Multi-Platform Multi-Modal 3D Grounding Dataset}
\label{sec:dataset}

\begin{figure}[t]
    \centering
    \includegraphics[width=\linewidth]{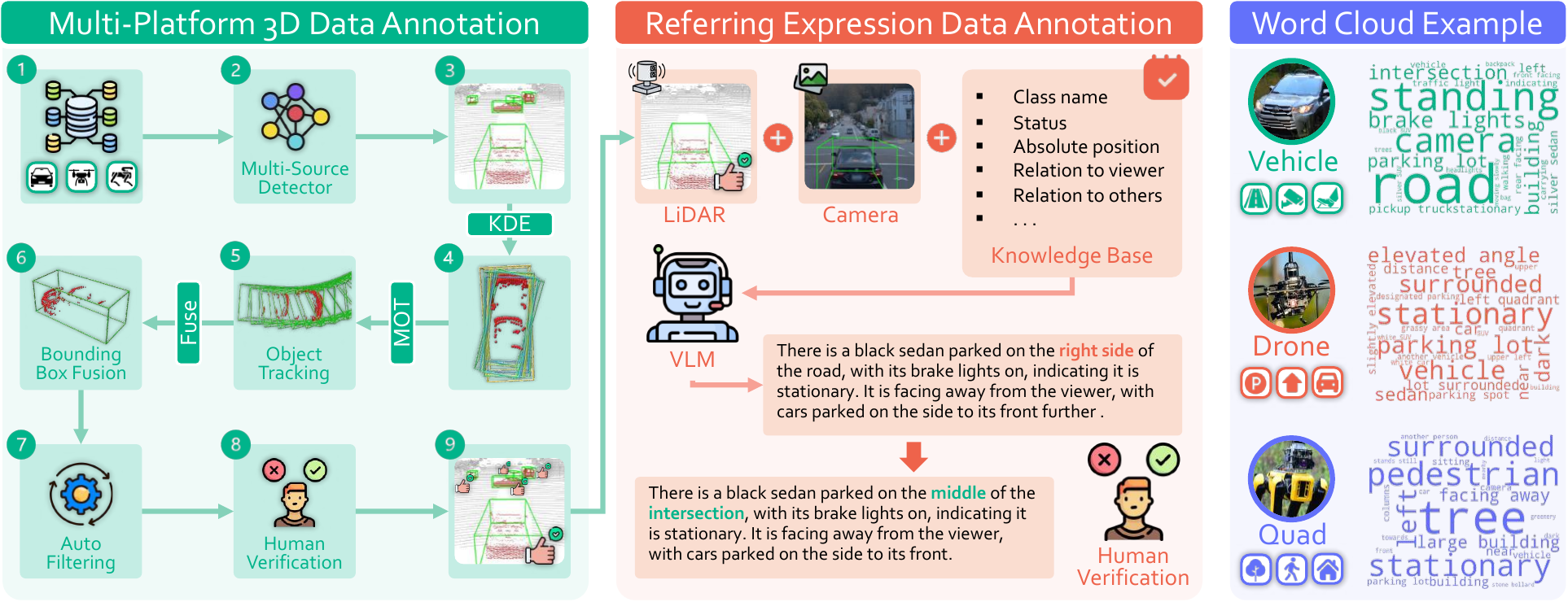}
    \vspace{-0.55cm}
    \caption{\textbf{Overview of annotation workflow.} 
    \textcolor{3eed_green}{\textbf{Left:}} We collect 3D boxes using multi-detector fusion, tracking, filtering, and manual verification across platforms. 
    \textcolor{3eed_red}{\textbf{Middle:}} Referring expressions are produced by prompting a VLM with structured cues (class, status, position, relations), followed by rule-based rewriting and human refinement. 
    \textcolor{3eed_blue}{\textbf{Right:}} Platform-specific word clouds highlight distinct linguistic patterns in descriptions across vehicle, drone, and quadruped agents.}
    \label{fig:annotation_pipeline}
\end{figure}

Existing 3D grounding datasets mainly target small, sensor‐fixed indoor spaces, leaving outdoor, multi-platform scenarios underexplored. To bridge this gap, we curate \ours, the first 3D grounding dataset that unifies data from {\includegraphics[width=0.021\linewidth]{figures/icons/vehicle.png}} \texttt{Vehicle}, {\includegraphics[width=0.028\linewidth]{figures/icons/drone.png}} \texttt{Drone}, and {\includegraphics[width=0.022\linewidth]{figures/icons/quadruped.png}} \texttt{Quadruped} platforms. We formalize the multi-modal, multi-platform 3D grounding task in Section~\ref{sec:formulation}, detail a two-stage annotation pipeline in Section~\ref{sec:annotation}, and present statistics that highlight the scale, diversity, and platform balance in Section~\ref{sec:analysis}.

\subsection{Task Formulation: 3D Grounding in the Wild}
\label{sec:formulation}

We define the multi-platform 3D grounding task in our dataset as $\mathcal{F}(\mathcal{P}^{\beta}, I^{\beta}, \mathcal{C}) \rightarrow \mathbf{b}^{\beta}$, where the model $\mathcal{F}$ maps input modalities, optionally including the point cloud $\mathcal{P}^{\beta} = \left\{ \mathbf{p}_{i} \right\}_{i=1}^{N^{\beta}}$, image $I^{\beta}$, and caption $\mathcal{C}$ to the corresponding 3D bounding box $\mathbf{b}^{\beta} \in \mathbb{R}^7$. Each point $\mathbf{p}_{i} = (p^x, p^y, p^z) \in \mathbb{R}^{3}$, and the bounding box is given by its center, dimensions, and orientation angle.  $\beta$ denotes the platform, including the {\includegraphics[width=0.021\linewidth]{figures/icons/vehicle.png}} \texttt{Vehicle}, {\includegraphics[width=0.03\linewidth]{figures/icons/drone.png}} \texttt{Drone}, and {\includegraphics[width=0.022\linewidth]{figures/icons/quadruped.png}} \texttt{Quadruped}, and $N^{\beta}$ is the number of point clouds for platform $\beta$. To precisely quantify spatial relationships, we also define the bird's-eye-view distance from \emph{target} to \emph{ego-platform} as $\rho$ and the relative pitch angle as $\theta^r$. In dataset curation and annotation, we explicitly consider \textbf{platform-specific factors} caused by inherent geometric differences.

\subsection{Dataset Curation \& Annotations}
\label{sec:annotation}
\noindent\textbf{Multi-Platform 3D Data Annotation.}
We collect the {\includegraphics[width=0.021\linewidth]{figures/icons/vehicle.png}} \texttt{Vehicle} sequences from Waymo~\cite{Sun_2020_CVPR},  and {\includegraphics[width=0.03\linewidth]{figures/icons/drone.png}} \texttt{Drone} and {\includegraphics[width=0.022\linewidth]{figures/icons/quadruped.png}} \texttt{Quadruped} sequences from M3ED~\cite{chaney2023m3ed}. We adopt a uniform \textbf{three-stage pipeline} for the \texttt{Drone}/\texttt{Quadruped} LiDAR–RGB (see Figure~\ref{fig:annotation_pipeline}, left).
\emph{\textbf{1)} Pseudo-label seeding:} State-of-the-art detectors~\cite{shi2020pv, shi2023pv, deng2021voxel, zhang2022not,yin2021center, yan2018second} trained on Waymo \cite{Sun_2020_CVPR}, nuScenes~\cite{caesar2020nuscenes}, and Lyft~\cite{houston2021one} produce platform-agnostic 3D boxes for every frame.
\emph{\textbf{2)} Automatic consolidation:} Kernel-density estimation (KDE) merges detector votes, a 3D multi-object tracker~\cite{fan2023once} enforces temporal coherence and fills missed detections, and the Tokenize-Anything~\cite{pan2024tokenize} model is used to project each box onto the RGB view to confirm its class; category conflicts are auto-flagged.
\emph{\textbf{3)} Human refinement:} Annotators polish the flagged boxes in the user interface, cross-validating to equalize accuracy across platforms. This hybrid scheme yields consistent annotations while limiting manual effort to roughly $100$s per frame.

\noindent\textbf{Referring Expression Data Annotation.} 
After collecting the 3D boxes, we attach platform-invariant language supervision through a parallel procedure (see Figure~\ref{fig:annotation_pipeline}, middle).  
\emph{\textbf{1)} Structured prompting:} Each 3D box is projected onto its RGB view, together with a knowledge base with five template slots \emph{category}, \emph{status}, \emph{absolute location}, \emph{egocentric position}, \emph{relation}, to a vision language model~\cite{qwen2-vl}. Few-shot expression examples in the prompt are used to guide the model to output a single, well-formed referring sentence. Platform-specific terms are normalized by {platform-invariant rewriting rules} to ensure consistent wording across vehicle, drone, and quadruped views.  
\emph{\textbf{2)} Human verification:} Annotators inspect the image, projected box, and caption in an interactive UI, checking semantic correctness, spatial fidelity, absence of ambiguity, and platform-consistency. Cases that are unsatisfactory will be discarded. This staged pipeline delivers concise, unambiguous expressions across vehicle, drone, and quadruped views, providing high-quality language targets for 3D visual grounding.

\begin{figure}[t]
    \centering
    \includegraphics[width=\linewidth]{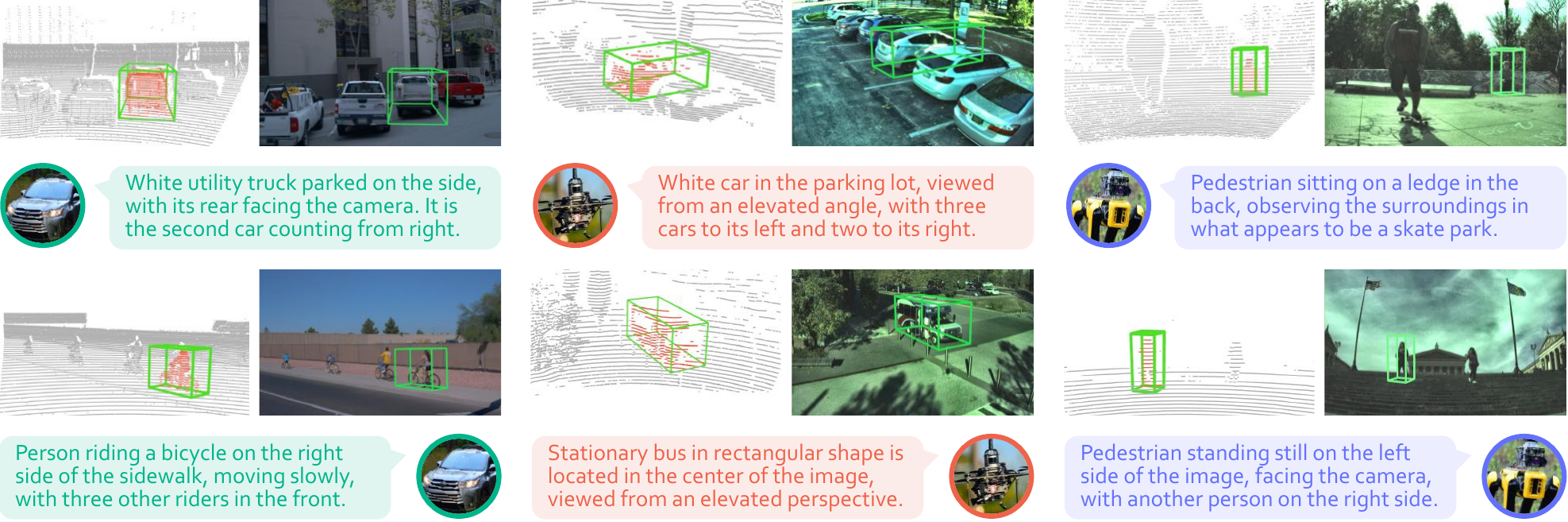}
    \caption{\textbf{Examples of multi-platform 3D grounding} from the \ours~dataset. There are clear discrepancies across both \emph{sensory data} (2D \& 3D) and \emph{referring expressions} from the {\includegraphics[width=0.021\linewidth]{figures/icons/vehicle.png}} \texttt{Vehicle}, {\includegraphics[width=0.03\linewidth]{figures/icons/drone.png}} \texttt{Drone}, and {\includegraphics[width=0.022\linewidth]{figures/icons/quadruped.png}} \texttt{Quadruped} platforms.}
    \label{fig:examples}
\end{figure}

\subsection{Dataset Statistics \& Analysis}
\label{sec:analysis}

\begin{figure}[t]
    \centering
    \includegraphics[width=\linewidth]{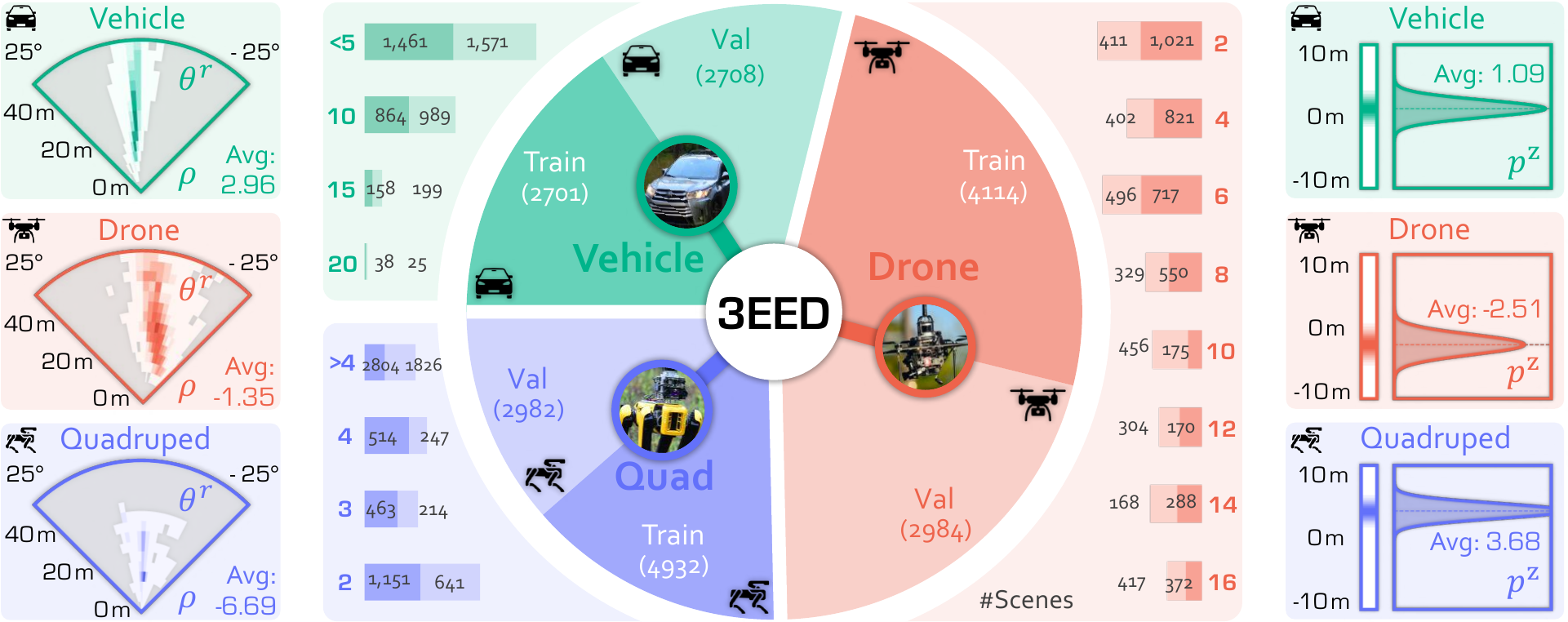}
    \caption{
    \textbf{Dataset statistics} of the three platforms in \ours.  
    \textbf{Left:} Target bounding box distributions in polar coordinates. Color intensity indicates the frequency of targets in each $(\rho, \theta^r)$ bin. 
    \textbf{Middle:} Scene distribution for train/val splits on each platform, along with per-scene object count histograms.  
    \textbf{Right:} Elevation distributions of input point cloud, $p^z$, reflecting view-dependent elevation biases.
    }
    \label{fig:statistics}
\end{figure}

\noindent\textbf{Benchmark Comparisons.} 
\ours{} is, to our knowledge, the \emph{first} outdoor {3D visual grounding} benchmark that {standardizes} sensing across three embodied platforms %
{\includegraphics[width=0.021\linewidth]{figures/icons/vehicle.png}}\;\texttt{Vehicle}, %
{\includegraphics[width=0.028\linewidth]{figures/icons/drone.png}}\;\texttt{Drone}, and %
{\includegraphics[width=0.022\linewidth]{figures/icons/quadruped.png}}\;\texttt{Quadruped} by using {synchronized LiDAR--RGB} acquisition.  As summarized in Table~\ref{tab:dataset_comparison}, our dataset provides $\mathbf{128{,}735}$ object bounding boxes and $\mathbf{22{,}439}$ human-verified referring expressions over $\mathbf{20{,}367}$ tightly time-aligned frames, focusing on the two safety-critical classes \emph{Vehicle} and \emph{Pedestrian}. Spatially, our scenes span up to $\mathbf{280\,m\,\times\,240\,m}$ horizontally and exceed $\mathbf{80\,m}$ in elevation, with an order of magnitude larger than any previous outdoor corpus, making it uniquely suited for studying long-range, cross-platform grounding. The train/val split is carefully balanced. As shown in Figure~\ref{fig:statistics} (middle), containing $2.7\text{k}/2.7\text{k}$ vehicle, $4.1\text{k}/2.9\text{k}$ drone, and $4.9\text{k}/2.9\text{k}$ quadruped scenes, enabling rigorous analysis of both platform-specific challenges and cross-platform generalization.

\noindent\textbf{Platform-Specific Analysis.}\,
To illuminate how \ours~supports \textbf{robust multi-platform downstream tasks}, we dissect the sensing geometry and scene composition of each agent in three dimensions:

\textbf{\emph{1)} Viewpoint geometry of targets:} 
Figure~\ref{fig:statistics}~(left) shows the distribution of pitch angle~$\theta^{r}$ and BEV range~$\rho$ for each 3D box. {\includegraphics[width=0.021\linewidth]{figures/icons/vehicle.png}} \texttt{Vehicle} data clusters at mid-range with near-zero pitch, typical of level driving. {\includegraphics[width=0.03\linewidth]{figures/icons/drone.png}} \texttt{Drone} covers larger~$\rho$ with steep negative~$\theta^{r}$ from top-down views. {\includegraphics[width=0.022\linewidth]{figures/icons/quadruped.png}} \texttt{Quadruped} stays close in~$\rho$ but varies widely in pitch due to ground-level perspective. These patterns expose models to varied spatial cues like ``behind'' and ``under'', improving generalization to novel viewpoints.

\textbf{\emph{2)} Per-platform object density:} 
Figure~\ref{fig:statistics} (middle) shows object density per platform. {\includegraphics[width=0.03\linewidth]{figures/icons/drone.png}} \texttt{Drone} captures the busiest scenes due to its wide view, {\includegraphics[width=0.021\linewidth]{figures/icons/vehicle.png}} \texttt{Vehicle} records moderate density, and {\includegraphics[width=0.022\linewidth]{figures/icons/quadruped.png}} \texttt{Quadruped} sees fewer but closer objects. This range enables 3EED to test the ability to disambiguate crowded scenes, maintain situational awareness, and localize small, nearby targets -- offering a challenging testbed for robust 3D grounding.

\textbf{\emph{3)} Input point-cloud geometry:} 
Figure~\ref{fig:statistics}~(right) shows the vertical distribution of LiDAR points $p^{z}$ per platform. {\includegraphics[width=0.021\linewidth]{figures/icons/vehicle.png}} \texttt{Vehicle} scans center around the sensor height, {\includegraphics[width=0.03\linewidth]{figures/icons/drone.png}} \texttt{Drone} captures top-down views, and {\includegraphics[width=0.022\linewidth]{figures/icons/quadruped.png}} \texttt{Quadruped} looks upward toward obstacles. These elevation biases affect how spatial terms like ``above'' or ``below'' are grounded, offering rich vertical language diversity across viewpoints.

\begin{figure}[t]
    \centering
    \includegraphics[width=\linewidth]{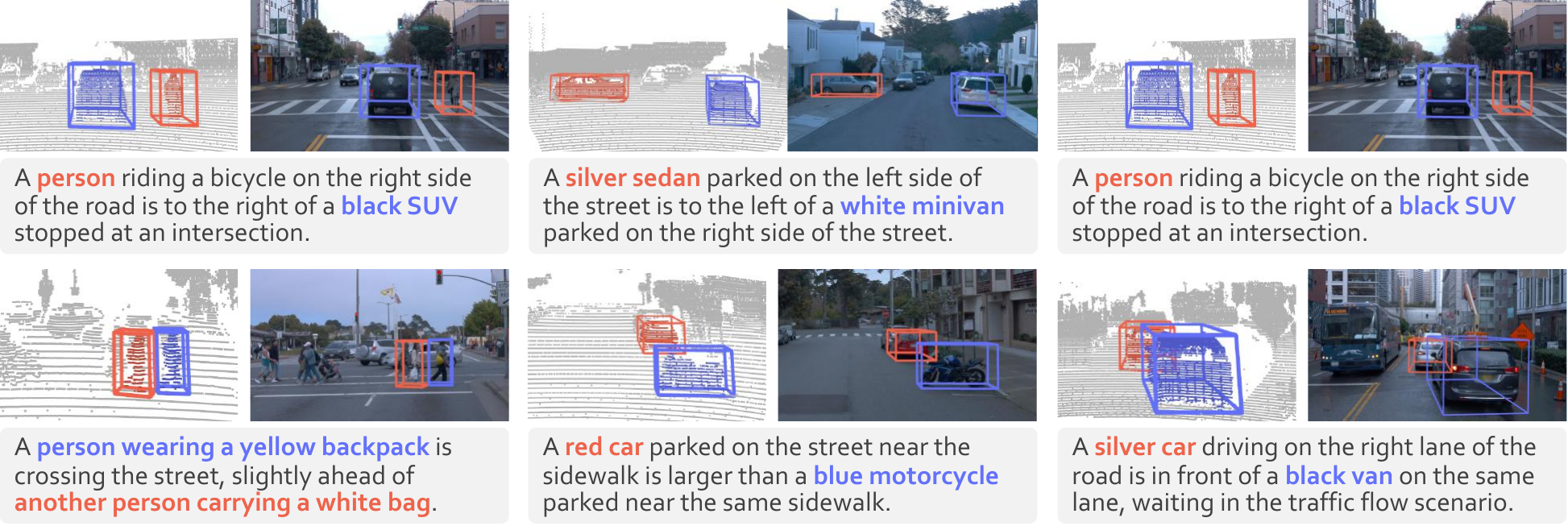}
    \vspace{-0.55cm}
    \caption{\textbf{Examples of multi-object 3D grounding} from the \ours~dataset.
    Given a scene and a multi-object expression, the goal of this task is to localize the 3D bounding box of each referred object by reasoning over both semantic attributes and inter-object spatial relationships.
    }
    \label{fig:examples_multi}
\end{figure}
\section{Benchmark Establishment}
\label{sec:method}

\subsection{Task Suite \& Evaluation Strategy}

The scale and heterogeneity of \ours, inclduing three embodied platforms, synchronized LiDAR--RGB sensing, and densely annotated outdoor scenes, enable a unified yet diagnostic suite of grounding benchmarks. We keep training schedules fixed across settings to make results comparable.

\emph{1) Single-platform, single-object grounding}: train and test on the same platform ({\includegraphics[width=0.021\linewidth]{figures/icons/vehicle.png}} \texttt{Vehicle} / {\includegraphics[width=0.03\linewidth]{figures/icons/drone.png}} \texttt{Drone} / {\includegraphics[width=0.022\linewidth]{figures/icons/quadruped.png}} \texttt{Quadruped}) to establish an in-domain reference under matched viewpoint and density statistics. This setting serves as a sanity check and a low-variance yardstick for later comparisons.

\emph{2) Cross-platform transfer}: To reflect real deployments where annotating Drone/Quadruped is costly, we adopt a zero-shot protocol: train on the data-rich {\includegraphics[width=0.021\linewidth]{figures/icons/vehicle.png}} \texttt{Vehicle} data and evaluate on the scarcer {\includegraphics[width=0.03\linewidth]{figures/icons/drone.png}} \texttt{Drone} and {\includegraphics[width=0.022\linewidth]{figures/icons/quadruped.png}} \texttt{Quadruped} data without target-domain supervision. All hyperparameters stay identical to the in-domain recipe; only the test platform changes. This isolates viewpoint/altitude/density shifts induced by different embodiments and measures cross-platform generalization.

\emph{3) Multi-object grounding}: A single query may describe multiple referents; the output must localize \emph{all} targets in the scene. We keep the same IoU thresholds as above but use \emph{joint correctness}: a query counts as correct iff every referred object is localized correctly. Figure~\ref{fig:examples_multi} illustrates several such scenes.

\emph{4) Multi-platform grounding}: Train on the union of all platforms and evaluate per-platform. We employ a balanced design across platforms while keeping the total training budget fixed. This tests whether pooled supervision can close transfer gaps without overfitting to platform-specific statistics.

\subsection{Challenges for Existing Methods}
Most 3D grounding models are designed for indoor RGB-D data, with dense, uniform points and small, consistent object sizes. On \ours, they face three key challenges:
\textbf{\emph{1)} Range-dependent sparsity:} LiDAR points thin out with distance, breaking indoor assumptions of dense neighborhoods.
\textbf{\emph{2)} Extreme scale variation:} Outdoor targets range from small cones to large vehicles, invalidating fixed-size anchors.
\textbf{\emph{3)} Cross-platform gaps:} Different viewpoints and sensor heights cause shifts in density and field of view unseen in indoor settings. As we will illustrate in the next section, these challenges reveal the need for outdoor- and platform-aware model designs.

\subsection{Unified Cross-Platform Baseline}

To kick-start research on \emph{cross-platform transfer} and \emph{multi-object grounding}, we present a scale-adaptive and agent-invariant baseline model tailored to \ours. It effectively addresses these challenges and serves as a strong reference point for future work in robust, general 3D visual grounding.

\noindent\textbf{Baseline Overview.}
We adapt previous work \cite{jain2022butd-detr} to our dataset: a scale-adaptive PointNet++ \cite{qi2017pointnet++} backbone encodes LiDAR, a frozen RoBERTa \cite{liu2019roberta} encodes language, and a Transformer predicts every referenced 3D box in one shot. Training blends box-regression, token-alignment, and contrastive multimodal losses. In the multi-object grounding setting, each target object is associated with a distinct positive map. We apply Hungarian matching to assign each query to a specific target object, enabling supervised learning via one-to-one loss computation.

\noindent\textbf{Cross-Platform Alignment (CPA).}
Before feature extraction, each scan is rotated to reduce roll and pitch so that gravity is consistently aligned with the global $z$-axis; drones additionally receive an altitude-normalizing height offset.
Placing all platforms in the same gravity-aligned frame reduces viewpoint- and elevation-induced discrepancies, so spatial relations such as ``above/below/behind'' are encoded in comparable coordinates across agents.
This simple, one-shot normalization lets the backbone spend capacity on object/content cues rather than pose correction, improving in-domain stability and yielding more reliable cross-platform generalization without any architecture change.

\noindent\textbf{Multi-Scale Sampling (MSS).}
Each PointNet++ layer queries neighborhoods at multiple radii from $0.6$\,m to $4.8$\,m, ensuring that the representation simultaneously preserves sharp local detail for nearby small objects and aggregates broad context for distant sparse targets.
This range-aware design avoids the failure modes of single-radius schemes (over-smoothing at close range, missing evidence at long range), directly countering LiDAR sparsity and extreme object-scale variation.
As a result, the encoder receives scale-complete evidence on all platforms, so it can localize both tiny traffic cones and large buses under diverse viewpoints and densities.
  
\noindent\textbf{Scale-Aware Fusion (SAF).}  
Features computed at all radii are fed to a lightweight MLP that produces dynamic, per-point weights and fuses the scales into a single embedding, emphasizing whichever radius best explains local geometry.
By adapting the contribution of fine vs.\ coarse context on the fly, SAF prevents “wrong-scale” decisions (\eg, using coarse features for small nearby objects or fine features for far sparse ones) and stabilizes predictions under large density shifts across platforms.
The module adds negligible parameters and latency while delivering scale-robust, agent-agnostic representations that complement CPA and MSS.

\section{Experiments}
\label{sec:experiments}

\begin{table}[t]
\centering
\caption{\textbf{Benchmark results of state-of-the-art models on \ours.}
Rows are grouped by the \emph{training platform}: \texttt{Vehicle} / \texttt{Drone} / \texttt{Quadruped} / \texttt{Union}, and columns report test performance on each platform; 
{\begingroup\setlength{\fboxsep}{0.6pt}\colorbox{3eed_gray!22}{diagonal cells}\endgroup} are \emph{in-domain}, while 
{\begingroup\setlength{\fboxsep}{0.6pt}\fcolorbox{darkgray}{white}{off-diagonals}\endgroup} are \emph{zero-shot cross-platform}.
The \textit{Platform Adaptation} column marks whether a method uses our platform-aware design (\cmark) or not (\xmark).
The \textit{Improve}~$\uparrow$ row in each block gives the absolute gain of \textit{Ours} over the strongest baseline under the same training protocol and metric.
All scores \texttt{Acc@25/50} are given in percentage (\%).
}
\vspace{-0.2cm}
\label{tab:single_obj_ground}
\resizebox{\linewidth}{!}{
\begin{tabular}{r|c|cc|cc|cc|cc}
    \toprule
    \multirow{2}{*}{\textbf{Method}}
    & \textbf{Platform}
    & \multicolumn{2}{c|}{{\includegraphics[width=0.021\linewidth]{figures/icons/vehicle.png}} \texttt{Vehicle}} 
    & \multicolumn{2}{c|}{{\includegraphics[width=0.03\linewidth]{figures/icons/drone.png}} \texttt{Drone}} & \multicolumn{2}{c|}{{\includegraphics[width=0.022\linewidth]{figures/icons/quadruped.png}} \texttt{Quadruped}} & \multicolumn{2}{c}{\textbf{Union}} 
    \\
    & \textbf{Adaptation}
    & \texttt{Acc@25} & \texttt{Acc@50}
    & \texttt{Acc@25} & \texttt{Acc@50}
    & \texttt{Acc@25} & \texttt{Acc@50} 
    & \texttt{Acc@25} & \texttt{Acc@50} 
    \\
    \midrule\midrule
    \rowcolor{3eed_green!15}\multicolumn{10}{l}{\textcolor{3eed_green}{$\bullet$~\textbf{Training Platform:}} {\includegraphics[width=0.021\linewidth]{figures/icons/vehicle.png}} \texttt{Vehicle}}
    \\
    BUTD\text{-}DETR~\cite{jain2022butd-detr} & \xmark & $52.38$ & $32.18$ & $1.54$ & $0.00$ & $10.18$ & $0.79$ & $23.70$ & $12.54$ 
    \\ 
    EDA~\cite{wu2023eda} & \xmark & $53.54$ & $34.87$ & $3.33$  & $0.05$ & $11.40$ & $0.62$ & $25.36$ & $13.81$ 
    \\ 
    WildRefer~\cite{lin2024wildrefer} & \xmark & $50.27$ & $9.85$ & $3.52$ & $0.34$ & $13.97$ & $3.76$ & $24.92$ & $5.12$ 
    \\
    \textbf{Ours} & \cmark & $78.37$ & $45.72$ & $18.16$ & $2.78$ & $36.04$ & $20.59$ & $45.93$ & $22.88$ 
    \\\midrule
    \textit{Improve} $\uparrow$ & - & \textcolor{3eed_green}{$\mathbf{+25.99}$} & \textcolor{3eed_green}{$\mathbf{+13.54}$} & \textcolor{3eed_green}{$\mathbf{+16.62}$} & \textcolor{3eed_green}{$\mathbf{+2.78}$} & \textcolor{3eed_green}{$\mathbf{+25.86}$} & \textcolor{3eed_green}{$\mathbf{+19.80}$} & \textcolor{3eed_green}{$\mathbf{+22.23}$} & \textcolor{3eed_green}{$\mathbf{+10.34}$}
    \\
    \midrule
    \rowcolor{3eed_red!15}\multicolumn{10}{l}{\textcolor{3eed_red}{$\bullet$~\textbf{Training Platform:}} {\includegraphics[width=0.03\linewidth]{figures/icons/drone.png}} \texttt{Drone}}
    \\
    BUTD\text{-}DETR~\cite{jain2022butd-detr} & \xmark & $15.08$ & $2.21$ & $40.85$ & $5.29$ & $6.90$ & $1.54$ & $20.55$ & $2.95$ 
    \\ 
    EDA~\cite{wu2023eda} & \xmark & $17.32$ & $4.81$ & $43.29$ & $7.10$ & $8.54$ & $2.71$ & $22.66$ & $4.88$
    \\ 
    WildRefer~\cite{lin2024wildrefer} & \xmark & $4.61$ & $0.69$ & $46.15$ & $8.21$ & $14.96$ & $5.40$ & $20.41$ & $4.52$ 
    \\
    \textbf{Ours} & \cmark & $29.01$ & $5.79$ & $47.55$ & $8.71$ & $31.32$ & $3.69$ & $34.56$ & $6.05$  
    \\\midrule
    \textit{Improve} $\uparrow$ & - & \textcolor{3eed_red}{$\mathbf{+13.93}$} & \textcolor{3eed_red}{$\mathbf{+3.58}$} & \textcolor{3eed_red}{$\mathbf{+6.70}$} & \textcolor{3eed_red}{$\mathbf{+3.42}$} & \textcolor{3eed_red}{$\mathbf{+24.42}$} & \textcolor{3eed_red}{$\mathbf{+2.15}$} & \textcolor{3eed_red}{$\mathbf{+14.01}$} &  \textcolor{3eed_red}{$\mathbf{+3.10}$}
    \\
    \midrule
    \rowcolor{3eed_blue!15}\multicolumn{10}{l}{\textcolor{3eed_blue}{$\bullet$~\textbf{Training Platform:}} {\includegraphics[width=0.022\linewidth]{figures/icons/quadruped.png}} \texttt{Quadruped}}
    \\
    BUTD\text{-}DETR~\cite{jain2022butd-detr} & \xmark & $14.76$ & $6.03$ & $9.92$ & $0.94$ & $32.38$ & $17.32$ & $18.59$ & $7.87$ 
    \\ 
    EDA~\cite{wu2023eda} & \xmark & $15.96$ & $6.83$ & $10.92$ & $1.44$ & $33.88$ & $18.52$ & $19.84$ & $8.70$ 
    \\ 
    WildRefer~\cite{lin2024wildrefer} & \xmark & $5.03$ & $0.87$ & $10.32$ & $0.84$ & $30.70$ & $19.59$ & $14.08$ & $6.54$ 
    \\
    \textbf{Ours} & \cmark & $20.52$ & $6.16$ & $10.52$ & $9.92$ & $35.69$ & $17.38$ & $21.43$ & $7.95$ 
    \\
    \textit{Improve} $\uparrow$ & - & \textcolor{3eed_blue}{$\mathbf{+5.76}$} & \textcolor{3eed_blue}{$\mathbf{+0.13}$} & \textcolor{3eed_blue}{$\mathbf{+0.60}$} & \textcolor{3eed_blue}{$\mathbf{+8.98}$} & \textcolor{3eed_blue}{$\mathbf{+3.31}$} & \textcolor{3eed_blue}{$\mathbf{+0.06}$} & \textcolor{3eed_blue}{$\mathbf{+2.84}$} & \textcolor{3eed_blue}{$\mathbf{+0.08}$}
    \\
    \midrule
    \rowcolor{3eed_gray!22}\multicolumn{10}{l}{\textcolor{darkgray}{$\bullet$~\textbf{Training Platform:}} \textbf{Union} ({\includegraphics[width=0.021\linewidth]{figures/icons/vehicle.png}} \texttt{Vehicle} + {\includegraphics[width=0.03\linewidth]{figures/icons/drone.png}} \texttt{Drone} + {\includegraphics[width=0.022\linewidth]{figures/icons/quadruped.png}} \texttt{Quadruped})}
    \\
    BUTD\text{-}DETR~\cite{jain2022butd-detr} & \xmark & $63.41$ & $40.88$ & $44.20$ & $8.28$ & $43.14$ & $20.94$ & $51.41$ & $24.80$ 
    \\ 
    EDA~\cite{wu2023eda} & \xmark & $65.50$ & $41.80$ & $46.00$ & $8.60$ & $44.00$ & $21.50$ & $52.46$ & $25.02$
    \\ 
    WildRefer~\cite{lin2024wildrefer} & \xmark & $51.51$ & $10.12$ & $50.27$ & $9.85$ & $45.36$ & $20.29$ & $49.27$ & $13.11$ 
    \\
    \textbf{Ours} & \cmark & $80.86$ & $50.11$ & $53.45$ & $9.75$ & $53.31$ & $24.08$ & $63.84$ & $29.66$ 
    \\\midrule
    \textit{Improve} $\uparrow$ & - & \textcolor{darkgray}{$\mathbf{+17.45}$} & \textcolor{darkgray}{$\mathbf{+9.23}$} & \textcolor{darkgray}{$\mathbf{+9.25}$} & \textcolor{darkgray}{$\mathbf{+1.47}$} & \textcolor{darkgray}{$\mathbf{+10.17}$} & \textcolor{darkgray}{$\mathbf{+3.14}$} & \textcolor{darkgray}{$\mathbf{+12.43}$} & \textcolor{darkgray}{$\mathbf{+4.86}$}
\\
    \bottomrule
\end{tabular}
}
\vspace{-0.2cm}
\end{table}

\subsection{Experimental Setups}

\noindent\textbf{Implementation Details.}
Our method is implemented in PyTorch, following the training schedule and optimization settings of previous work~\cite{jain2022butd-detr}, but optimized for efficiency. Raw LiDAR from any platform is uniformly down-sampled to $16{,}384$ points and encoded by a PointNet++ backbone~\cite{qi2017pointnet++} trained from scratch; its final layer yields $1{,}024$ visual tokens. An MLP assigns each token an objectness score, and the top $256$ tokens are input into a six-layer Transformer decoder. Objectness is supervised with focal loss by labeling the four nearest points to every ground-truth center as positives. We freeze RoBERTa, use a learning rate of $1 \times 10^{-3}$ for the visual encoder and $1 \times 10^{-4}$ for all other layers, and train for $100$ epochs on two NVIDIA RTX 4090 GPUs. See \textbf{Appendix} for more details.

\noindent\textbf{Evaluation Metrics.}
Following~\cite{chen2020scanrefer, achlioptas2020referit3d, lin2024wildrefer}, we report \emph{Top-1 Acc}, counting a success when the top box exceeds a chosen IoU. We evaluate at \texttt{Acc@25} (lenient) and \texttt{Acc@50} (strict), and report mean IoU (\texttt{mIoU}) for overall quality. In multi-object setup, all objects must meet the IoU threshold, penalizing misses and false positives. Results are averaged over official train/val splits for fair comparison.

\noindent\textbf{Baselines.}
We adapt two representative baselines. \emph{EDA} \cite{wu2023eda} is a prior art on indoor datasets by decoupling sentences into object, attribute, relation, and pronoun tokens, enforcing dense token-point alignment. However, it relies on dense scenes and grammar-consistent text, making it fragile under sparse LiDAR, large object-size variation, and diverse viewpoints. \emph{BUTD-DETR} \cite{jain2022butd-detr} uses a DETR-style decoder~\cite{carion2020end} with ScanNet box proposals and synthetic prompts but struggles on drone and quadruped data due to its dependence on indoor detectors. Neither baseline addresses range-dependent sparsity, scale variation, or cross-platform biases, motivating our scale-adaptive, agent-invariant baseline. Due to space limits, additional details are provided in the \textbf{Appendix}.

\begin{table}[t]
\centering
\caption{\textbf{Benchmark results of state-of-the-art models} on the \ours~dataset. 
The performances are measured under the \textit{multi-object} setting on the {\includegraphics[width=0.021\linewidth]{figures/icons/vehicle.png}} \texttt{Vehicle} platform.
We report the class-wise performance on \texttt{Acc@25}, \texttt{Acc@50}, and \texttt{mIoU} metrics. All scores are given in percentage (\%).}
\vspace{-0.2cm}
\label{tab:multi_obj_ground}
\resizebox{\linewidth}{!}{
\begin{tabular}{r|ccc|ccc|ccc}
    \toprule
    \multirow{2}{*}{\textbf{Method}} & \multicolumn{3}{c|}{\textbf{Car}} & \multicolumn{3}{c|}{\textbf{Pedestrian}} & \multicolumn{3}{c}{\textbf{Average}} 
    \\
    & ~\texttt{Acc@25}~ & ~\texttt{Acc{@}50}~ & \texttt{mIoU} & ~\texttt{Acc@25}~ & ~\texttt{Acc@50}~ & \texttt{mIoU} & ~\texttt{Acc@25}~ & ~\texttt{Acc@50}~ & \texttt{mIoU} 
    \\
    \midrule\midrule
    BUTD\text{-}DETR~\cite{jain2022butd-detr} & $30.92$ & $19.83$ &  $52.39$ & $26.56$ & $18.75$ & $37.28$  & $25.40$& $17.91$&  $47.88$
    \\
    EDA~\cite{wu2023eda} & $29.58$ & $26.21$ & $56.73$ & $28.15$ & $14.75$ &  $38.37$ & $26.91$& $25.92$ & $51.07$
    \\
    \textbf{Ours} & $37.21$ & $33.14$ & $59.28$ & $32.81$ & $20.31$ & $54.21$  & $32.32$& $29.89$ & $56.40$
    \\\midrule
    \textit{Improve} $\uparrow$ & \textcolor{3eed_green}{$\mathbf{+7.63}$} & \textcolor{3eed_green}{$\mathbf{+14.63}$} & \textcolor{3eed_green}{$\mathbf{+6.89}$}  & \textcolor{3eed_green}{$\mathbf{+4.66}$} & \textcolor{3eed_green}{$\mathbf{+1.56}$} & \textcolor{3eed_green}{$\mathbf{+15.84}$} & \textcolor{3eed_green}{$\mathbf{+5.41}$} & \textcolor{3eed_green}{$\mathbf{+3.97}$} & \textcolor{3eed_green}{$\mathbf{+5.33}$} 
    \\
    \bottomrule
\end{tabular}
}
\end{table}

\begin{table}[t]
    \centering
    \begin{minipage}[t]{0.485\textwidth}
    \centering
    \caption{\textbf{Ablation study on components.}  
    Multi-platform results (\texttt{Acc@25/50}, \%) comparing \textit{Full} \vs~removing one module (\textit{--CPA}, \textit{--MSS}, \textit{--SAF}).
    Dropping any module degrades performance, showing their complementarity.
    }
    \vspace{-0.2cm}
    \label{tab:abl_components}
    \resizebox{\textwidth}{!}{
    \begin{tabular}{l|p{22pt}<{\centering}p{27pt}<{\centering}|p{22pt}<{\centering}p{27pt}<{\centering}|p{22pt}<{\centering}p{27pt}<{\centering}}
    \toprule
    \multirow{2}{*}{\textbf{Method}} & \multicolumn{2}{c|}{{\includegraphics[width=0.042\linewidth]{figures/icons/vehicle.png}} \texttt{Vehicle}} 
    & \multicolumn{2}{c|}{{\includegraphics[width=0.06\linewidth]{figures/icons/drone.png}} \texttt{Drone}} & \multicolumn{2}{c}{{\includegraphics[width=0.044\linewidth]{figures/icons/quadruped.png}} \texttt{Quadruped}}
    \\
    & \texttt{Acc@25} & \texttt{Acc@50} & \texttt{Acc@25} & \texttt{Acc@50} & \texttt{Acc@25} & \texttt{Acc@50}
    \\\midrule\midrule
    $-$ \textbf{CPA} & $71.76$ & $50.42$ & $51.84$ & $9.32$ & $49.93$ & $23.53$ 
    \\
    $-$ \textbf{MSS} & $75.65$ & $45.98$ & $46.85$ & $8.51$ & $51.40$ & $24.25$ 
    \\
    $-$ \textbf{SAF} & $80.38$ & $50.03$  & $52.25$ & $10.19$ & $51.98$ & $24.80$
    \\
    \textbf{Full}  & $80.86$ & $50.11$ & $53.45$ & $9.75$ & $53.31$ & $24.08$
    \\
    \bottomrule
    \end{tabular} 
    }
    \end{minipage}
    \hfill
    \begin{minipage}[t]{0.485\textwidth}
    \centering
    \caption{\textbf{Ablation study on scene complexity.} 
    Results (\texttt{Acc@25/50}, \%) with scenes grouped by the number of objects per scene ($1$–$3$, $4$–$6$, $7$–$9$, $>9$).
    The performances are measured under the \textit{multi-platform} setting.}
    \vspace{-0.2cm}
    \label{tab:abl_stats}
    \resizebox{\textwidth}{!}{
    \begin{tabular}{l|p{22pt}<{\centering}p{27pt}<{\centering}|p{22pt}<{\centering}p{27pt}<{\centering}|p{22pt}<{\centering}p{27pt}<{\centering}}
    \toprule
    \textbf{Object} & \multicolumn{2}{c|}{{\includegraphics[width=0.042\linewidth]{figures/icons/vehicle.png}} \texttt{Vehicle}} 
    & \multicolumn{2}{c|}{{\includegraphics[width=0.06\linewidth]{figures/icons/drone.png}} \texttt{Drone}} & \multicolumn{2}{c}{{\includegraphics[width=0.044\linewidth]{figures/icons/quadruped.png}} \texttt{Quadruped}}
    \\
    \textbf{Count} & \texttt{Acc@25} & \texttt{Acc@50}
    & \texttt{Acc@25} & \texttt{Acc@50}
    & \texttt{Acc@25} & \texttt{Acc@50}
    \\\midrule\midrule
    $1 - 3$ & $62.24$ & $36.20$ & $52.07$ & $25.06$ & $71.23$ & $61.75$ 
    \\
    $4 - 6$ & $60.86$ & $44.00$ & $53.95$ & $10.76$ & $43.53$ & $9.48$ 
    \\
    $7 - 9$ & $59.55$ & $35.73$ & $53.44$ & $2.23$ & $30.75$ & $5.47$ 
    \\
    $>9$ & $63.39$ & $50.45$ & $31.82$ & $4.09$ & $25.15$ & $0.83$ 
    \\
    \bottomrule
    \end{tabular}
    }
    \end{minipage}
\end{table}

\subsection{Comparative Study}

\noindent\textbf{Cross-Platform Generalization.} Table~\ref{tab:single_obj_ground} compares existing 3D grounding backbones under in-distribution (\emph{single‐platform}) and out‐of‐distribution (\emph{cross‐platform}) settings. 

\emph{1) Single-Platform vs. Cross-Platform.} When trained on {\includegraphics[width=0.021\linewidth]{figures/icons/vehicle.png}} \texttt{Vehicle} data, BUTD-DETR \cite{jain2022butd-detr} achieves \texttt{Acc@25} of $52.38$ on the vehicle test split, but drops to $1.54$ on drone and $10.18$ on quadruped, exposing severe generalization gaps due to differing viewpoints, object scales, and LiDAR densities. 

\emph{2) Cross-Platform Transfer Gains.} Our scale-adaptive backbone with platform alignment substantially narrows this gap. For example, training on {\includegraphics[width=0.03\linewidth]{figures/icons/drone.png}} \texttt{Drone} and evaluating on {\includegraphics[width=0.021\linewidth]{figures/icons/vehicle.png}} \texttt{Vehicle} boosts \texttt{Acc@25} by $+13.93$ over the baseline, demonstrating stronger transfer from aerial to ground perspectives. 

\emph{3) Unified Multi-Platform Training.} A unified model trained jointly on all three platforms delivers balanced performance, with \texttt{Acc@25} of $63.84$, $53.45$, and $53.31$ on vehicle, drone, and quadruped, respectively, yielding an average gain of $+12.29$ over the best method. This confirms the critical role of \ours~in providing diverse supervision for building truly generalizable 3D grounding systems.

\noindent\textbf{Coherent Object Co-grounding.}
Table~\ref{tab:multi_obj_ground} presents the evaluation results on our dataset for the \emph{multi-object grounding} task. Notably, in this setting, \texttt{Acc@25} is a strict metric that requires all objects mentioned in the description to be correctly grounded, while \texttt{mIoU} captures the average IoU across individual predicted-ground truth pairs.  Existing methods such as BUTD-DETR achieve moderate \texttt{mIoU} ($47.88$) but low joint grounding ($\texttt{Acc@25}=25.40$), revealing their tendency to localize objects in isolation rather than reason about them collectively. In contrast, our baseline leverages multi-scale sampling and dynamic feature fusion to build discriminative representations that capture both fine details and broad context, essential for disambiguating multiple objects of varying size and distance. These design choices deliver substantial improvements in both metrics, demonstrating markedly stronger multi-object reasoning and tighter language-to-3D alignment in complex outdoor scenes.

\begin{figure}[t]
    \centering
    \includegraphics[width=\linewidth]{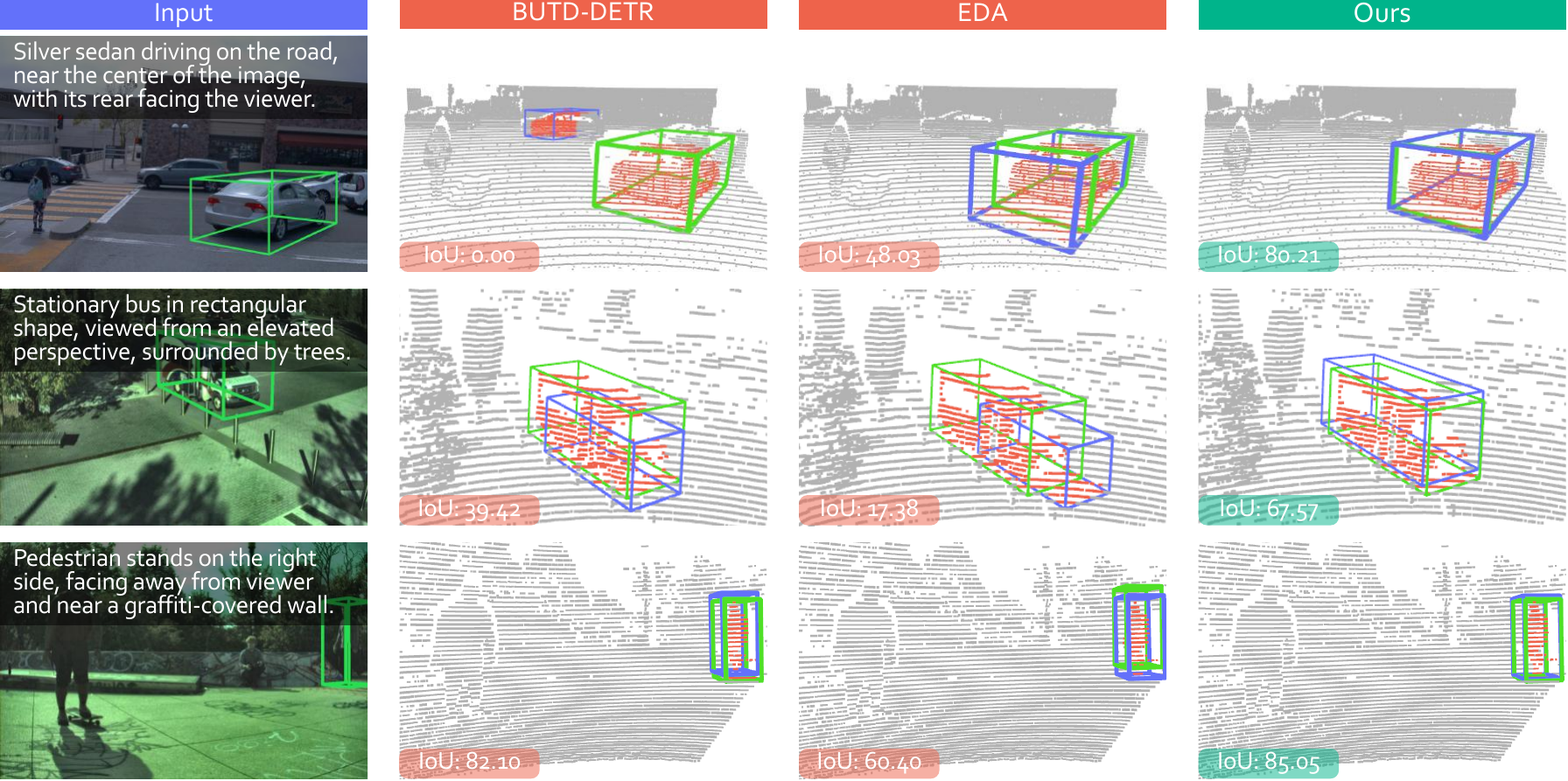}
    \vspace{-0.55cm}
    \caption{\textbf{Qualitative comparisons} of 3D grounding approaches on the \ours~dataset. We show the comparisons under the \emph{multi-platform} setting. The three examples are from the {\includegraphics[width=0.021\linewidth]{figures/icons/vehicle.png}} \texttt{Vehicle}, {\includegraphics[width=0.03\linewidth]{figures/icons/drone.png}} \texttt{Drone}, and {\includegraphics[width=0.022\linewidth]{figures/icons/quadruped.png}} \texttt{Quadruped} platforms, respectively. Kindly refer to the appendix for additional results.}
    \label{fig:qualitative}
\end{figure}

\noindent\textbf{Qualitative Assessments.}
Figure~\ref{fig:qualitative} showcases representative \emph{multi-platform grounding} results on vehicle, drone, and quadruped data. Our unified model consistently outputs precise, tightly aligned 3D boxes despite drastic shifts in viewpoint, object scale, and point-cloud density. In contrast, baseline methods like BUTD-DETR \cite{jain2022butd-detr} and EDA \cite{wu2023eda} often yield misaligned or fragmented predictions, especially under challenging aerial and low-angle quadruped perspectives. These comparisons underscore our ability to learn genuine cross-platform invariance and deliver reliable grounding across diverse embodied sensing scenarios.

\subsection{Ablation Study}
\noindent\textbf{Component Analysis.}
Table~\ref{tab:abl_components} shows that our modules target different sources of error and, together, improve both in-domain accuracy and cross-platform transfer.

\textbf{\emph{1) CPA (Cross-Platform Alignment)}}
is the primary driver by rotating each scene to cancel roll and pitch and normalizing the height offset to reduce elevation bias, it effectively maps data into a gravity-aligned frame with comparable coordinates across platforms.
This substantially reduces viewpoint-induced discrepancies (\eg, for ``above/below/behind'') so the backbone need not spend capacity correcting pose biases.
Consequently, CPA yields a large {\includegraphics[width=0.021\linewidth]{figures/icons/vehicle.png}} \texttt{Vehicle} gain from $71.76$ to $80.86 (+9.10)$ in $\texttt{Acc@25}$ in-domain and leads to more stable cross-platform transfer.

\textbf{\emph{2) MSS (Multi-Scale Sampling)}} addresses the core failure of single-radius neighborhoods under range-dependent sparsity. A small radius preserves nearby details but fails at long range (no points in the neighborhood), whereas a large radius recovers distant context but over-smooths close objects. MSS samples a \emph{wide spectrum of radii} per query, so each point receives both \emph{fine-detail} evidence (for near objects) and \emph{global-context} evidence (for distant targets). This directly improves in-domain accuracy by recovering long-range evidence while avoiding close-range over-smoothing (reflected by the 
$+5.21$ $\texttt{Acc@25}$ gain in {\includegraphics[width=0.021\linewidth]{figures/icons/vehicle.png}} \texttt{Vehicle}), and it improves cross-platform transfer because receptive-field behavior no longer depends on platform-specific altitude/FoV statistics: the same multi-radius coverage remains valid on sparser Drone views, narrowing cross-platform gaps.

\textbf{\emph{3) SAF (Scale-Aware Fusion)}} then learns per-point weights over scales via a lightweight MLP, ensuring the model \emph{uses the right scale at the right place}; this stabilizes predictions under density shifts (\eg, {\includegraphics[width=0.022\linewidth]{figures/icons/quadruped.png}} \texttt{Quadruped} $\texttt{Acc@25}$ $51.98$ to $53.31$), and further improves transfer by preventing a single fixed scale from dominating when switching platforms.

Combined, the modules deliver the best overall results: {\includegraphics[width=0.021\linewidth]{figures/icons/vehicle.png}} \texttt{Vehicle} ($80.86/50.11$), {\includegraphics[width=0.03\linewidth]{figures/icons/drone.png}} \texttt{Drone} ($53.45/9.75$), and {\includegraphics[width=0.022\linewidth]{figures/icons/quadruped.png}} \texttt{Quadruped} ($53.31/24.08$) ($\texttt{Acc@25/50}$), confirming that \emph{CPA handles cross-platform alignment, MSS provides evidence coverage, and SAF enforces adaptive selection}.

\begin{table}[t]
    \centering
    \caption{\textbf{Platform statistics and cross-platform performance.}
    Left: dataset statistics—average \emph{annotated objects per scene} and \emph{LiDAR points per object} (counts).
    Right: cross-evaluation matrix with rows as the \emph{training} platform and columns as the \emph{test} platform (diagonal = in-domain; off-diagonal = zero-shot). 
    Metrics are \texttt{Acc@25/50} in \% (\texttt{IoU 0.25/0.50}).}
    \vspace{-0.2cm}
    \resizebox{\linewidth}{!}{\begin{tabular}{c|c|c|cc|cc|cc}
    \toprule
    \multirow{2}{*}{\textbf{Platform}} & \textbf{Average} & \textbf{Average} & \multicolumn{2}{c|}{{\includegraphics[width=0.021\linewidth]{figures/icons/vehicle.png}} \texttt{Vehicle}} 
    & \multicolumn{2}{c|}{{\includegraphics[width=0.03\linewidth]{figures/icons/drone.png}} \texttt{Drone}} & \multicolumn{2}{c}{{\includegraphics[width=0.022\linewidth]{figures/icons/quadruped.png}} \texttt{Quadruped}}
    \\
    & \textbf{\#Objects / Scene} & \textbf{\#Points / Object} & \texttt{Acc@25} & \texttt{Acc@50} & \texttt{Acc@25} & \texttt{Acc@50} & \texttt{Acc@25} & \texttt{Acc@50}
    \\
    \midrule\midrule
    {\includegraphics[width=0.021\linewidth]{figures/icons/vehicle.png}} \texttt{Vehicle} & $4.77$ & $\mathbf{462.89}$ & \cellcolor{3eed_green!15}$78.37$ & \cellcolor{3eed_green!15}$45.72$ & $18.16$ & $2.78$ & $36.04$ & $20.59$ 
    \\
    {\includegraphics[width=0.03\linewidth]{figures/icons/drone.png}} \texttt{Drone} & $\mathbf{8.05}$ & $102.24$ & $29.01$ & $5.79$ & \cellcolor{3eed_red!15}$47.55$ & \cellcolor{3eed_red!15}$8.71$ & $31.32$ & $3.69$ 
    \\
    {\includegraphics[width=0.022\linewidth]{figures/icons/quadruped.png}} \texttt{Quadruped} & $5.83$ & $112.17$ & $20.52$ & $6.16$ & $10.52$ & $9.92$ & \cellcolor{3eed_blue!15}$35.69$ & \cellcolor{3eed_blue!15}$17.38$ 
    \\
    \bottomrule
    \end{tabular}}
    \label{tab:ablation_platform}
\end{table}

\noindent\textbf{Object Density Impact.}
We analyze how referential grounding performance varies with the object density per scene. We divide test samples into bins based on the number of annotated 3D bounding boxes ($1$-$3$, $4$–$6$, $7$–$9$, $10$+), and compute the average \texttt{Acc@25} for each bin. As shown in Table~\ref{tab:abl_stats}, accuracy consistently drops as object count increases. On the {\includegraphics[width=0.022\linewidth]{figures/icons/quadruped.png}} \texttt{Quadruped} platform, \texttt{Acc@25} drops from $71.23$ in scenes with $1$–$3$ objects to $30.75$ in scenes with $7$–$9$ objects. This reflects the increased difficulty of resolving referential ambiguity in cluttered environments.

\noindent\textbf{Platform Complexity Impact.}  
Table~\ref{tab:ablation_platform} breaks down grounding performance by platform alongside two key scene statistics: mean LiDAR points per object and mean object count per scene. {\includegraphics[width=0.03\linewidth]{figures/icons/drone.png}} \texttt{Drone} scenes suffer the lowest \texttt{Acc@50}, driven by extreme sparsity (just $102$ points/object \vs $462$ for {\includegraphics[width=0.021\linewidth]{figures/icons/vehicle.png}} \texttt{Vehicle} and $112$ for {\includegraphics[width=0.022\linewidth]{figures/icons/quadruped.png}} \texttt{Quadruped}) and the highest object density ($8.05$ objects/scene), which together amplify distractors and hinder precise localization. Quadruped data, with moderate density ($112$ points/object) but fewer objects, sits between drone and vehicle performance. These disparities, including ultra-sparse returns and elevated clutter, explain the pronounced aerial performance gap.  

\section{Conclusion}
\label{sec:conclusion}

We introduced \ours, a large-scale, multi-platform, multi-modal benchmark for outdoor 3D visual grounding, featuring $128{,}000$ objects and $22{,}000$ expressions, which is $10\times$ larger than existing datasets. We proposed scalable annotation, platform-aware normalization, and cross-modal alignment to support robust grounding. Our benchmark reveals cross-platform performance gaps, highlighting challenges for generalizable 3D grounding. We release our dataset and baseline models, hoping to advance the future development of language-driven embodied 3D perception.

\vspace{0.2cm}
\beginappendix

\appendix
\startcontents[appendices]
\printcontents[appendices]{l}{1}{\setcounter{tocdepth}{2}}
\vspace{0.3cm}

\section{The 3EED Dataset}

In this section, we provide a comprehensive overview of the \ours~dataset, including its motivation, collection methodology, and unique characteristics. We describe the design choices made to ensure diversity in sensor platforms, scene composition, and language annotation, and highlight the potential to support research in 3D visual grounding across real-world embodied platforms.

\subsection{Overview}

Our dataset is built on top of two existing autonomous driving and robotics datasets: \textbf{Waymo Open Dataset}~\cite{Sun_2020_CVPR} and \textbf{M3ED}~\cite{chaney2023m3ed}. Our dataset includes point cloud and image data collected from three distinct embodied platforms -- {\includegraphics[width=0.021\linewidth]{figures/icons/vehicle.png}} \texttt{Vehicle}, {\includegraphics[width=0.028\linewidth]{figures/icons/drone.png}} \texttt{Drone}, and {\includegraphics[width=0.022\linewidth]{figures/icons/quadruped.png}} \texttt{Quadruped} -- capturing scenes from street-level, aerial, and low-ground perspectives, respectively. The referring expressions are generated by Qwen2-VL-72B~\cite{qwen2-vl}, covering five aspects: \emph{category}, \emph{status}, \emph{absolute location}, \emph{egocentric position}, and \emph{spatial relation}, with human verification.

The full dataset contains $20{,}367$ multi-modal scenes, $22{,}439$ referring expressions, and $128{,}735$ annotated 3D object instances across three sensor platforms. The \textbf{training set} consists of $11{,}747$ scenes, with $12{,}733$ captions and $70{,}062$ objects, while the \textbf{validation set} includes $8{,}620$ scenes, $9{,}706$ captions, and ${58,}691$ objects. 

Breaking down by platform: the {\includegraphics[width=0.021\linewidth]{figures/icons/vehicle.png}} \texttt{Vehicle} split provides $5{,}409$ scenes and $25{,}818$ objects; the {\includegraphics[width=0.022\linewidth]{figures/icons/quadruped.png}} \texttt{Quadruped} split includes $7{,}860$ scenes and $45{,}797$ objects; and the {\includegraphics[width=0.028\linewidth]{figures/icons/drone.png}} \texttt{Drone} split contributes the portion with $7{,}098$ scenes and $57{,}138$ objects. This distribution reflects the platform diversity and scale of our dataset, supporting cross-platform and cross-viewpoint grounding evaluation.

This cross-platform, cross-viewpoint composition allows our dataset to serve as a unified benchmark for 3D grounding under varying spatial configurations, sensor geometries, and linguistic descriptions. It enables the evaluation of platform-agnostic language understanding in real-world conditions.

\subsection{Dataset Curation Details}

This section details the data sourcing, 3D bounding box annotation pipeline, and referring expression generation process used to construct the \ours~dataset. We describe how annotated 3D boxes are curated across platforms using a combination of pretrained detectors, tracking, and manual refinement, and how language expressions are generated and verified to ensure grounding quality and consistency across scenes.

\begin{table}[t]
\centering
\caption{Statistics of the \ours~dataset across platforms and splits.}
\label{tab:3eed_stats}
\vspace{-0.2cm}
\renewcommand{\arraystretch}{1.2}
\setlength{\tabcolsep}{15pt}
\begin{tabular}{l|ccc}
\toprule
\textbf{Platform} & \textbf{\# Scenes} & \textbf{\# Captions} & \textbf{\# Objects} \\
\midrule\midrule
\rowcolor{3eed_green!15}\multicolumn{4}{l}{\textcolor{3eed_green}{\textit{\textbf{Training}}} }\\
\quad {\includegraphics[width=0.021\linewidth]{figures/icons/vehicle.png}} \texttt{Vehicle}         & $2{,}701$ & $3{,}687$ & $12{,}790$ \\
\quad {\includegraphics[width=0.028\linewidth]{figures/icons/drone.png}} \texttt{Drone}      & $4{,}114$ & $4{,}114$ & $30{,}222$ \\
\quad {\includegraphics[width=0.022\linewidth]{figures/icons/quadruped.png}} \texttt{Quadruped}   & $4{,}932$ & $4{,}932$ & $27{,}050$ \\
\quad \textbf{Total }  & $\mathbf{11{,}747}$ & $\mathbf{12{,}733}$ & $\mathbf{70{,}062}$ \\
\midrule
\rowcolor{3eed_green!15}\multicolumn{4}{l}{\textcolor{3eed_green}{\textit{\textbf{Validation}}}} \\
\quad {\includegraphics[width=0.021\linewidth]{figures/icons/vehicle.png}} \texttt{Vehicle}         & $2{,}708$ & $3{,}794$ & $13{,}082$ \\
\quad {\includegraphics[width=0.028\linewidth]{figures/icons/drone.png}} \texttt{Drone}      & $2{,}984$ & $2{,}984$ & $26{,}916$ \\
\quad {\includegraphics[width=0.022\linewidth]{figures/icons/quadruped.png}} \texttt{Quadruped}   & $2{,}928$ & $2{,}928$ & $18{,}748$ \\
\quad \textbf{Total}    & $\mathbf{8{,}620}$ & $\mathbf{9{,}706}$ & $\mathbf{58{,}691}$ \\
\midrule
\rowcolor{darkgray!15} \textbf{Summary}          & $\mathbf{20{,}367}$ & $\mathbf{22{,}439}$ & $\mathbf{128{,}735}$ \\
\bottomrule
\end{tabular}
\end{table}

\subsubsection{Data Sources}

The dataset is built on top of two large-scale real-world 3D perception datasets: \textbf{Waymo Open Dataset}~\cite{Sun_2020_CVPR} and \textbf{M3ED}~\cite{chaney2023m3ed}.

\textbf{Waymo Open Dataset}~\cite{Sun_2020_CVPR} provides high-resolution LiDAR and RGB data collected from vehicle-mounted sensors in urban and suburban driving environments. We use a subset of Waymo annotated scenes to construct the {\includegraphics[width=0.021\linewidth]{figures/icons/vehicle.png}} \texttt{Vehicle} portion of our dataset, leveraging its high-quality 3D bounding boxes as ground truth. Our annotations are built independently on top of their publicly available sequences.

\textbf{M3ED Dataset}~\cite{chaney2023m3ed} is a multi-platform dataset, featuring synchronized RGB and LiDAR streams from both quadruped robots and aerial drones operating in various outdoor scenes. The {\includegraphics[width=0.028\linewidth]{figures/icons/drone.png}} \texttt{Drone} and {\includegraphics[width=0.022\linewidth]{figures/icons/quadruped.png}} \texttt{Quadruped} portions of our dataset are derived from M3ED. Since M3ED does not contain pre-annotated 3D bounding boxes, we adopt a semi-automatic annotation pipeline that combines multiple pretrained detectors, trajectory tracking, and human refinement to generate high-quality 3D boxes.

\subsubsection{Annotation Details on 3D Bounding Boxes}

The 3D bounding box annotations in \ours~are obtained through a combination of high-quality existing labels and a carefully designed cross-platform annotation pipeline.

\noindent\textbf{Vehicle Platform.} For the {\includegraphics[width=0.021\linewidth]{figures/icons/vehicle.png}} \texttt{Vehicle} platform, we adopt 3D object annotations directly from the official Waymo Open Dataset~\cite{Sun_2020_CVPR}, which provides dense, high-accuracy bounding boxes for traffic participants such as vehicles, pedestrians, and cyclists etc.. These annotations are widely regarded as reliable and are used without further modification.

\begin{figure}[t]
    \centering
    \includegraphics[width=\linewidth]{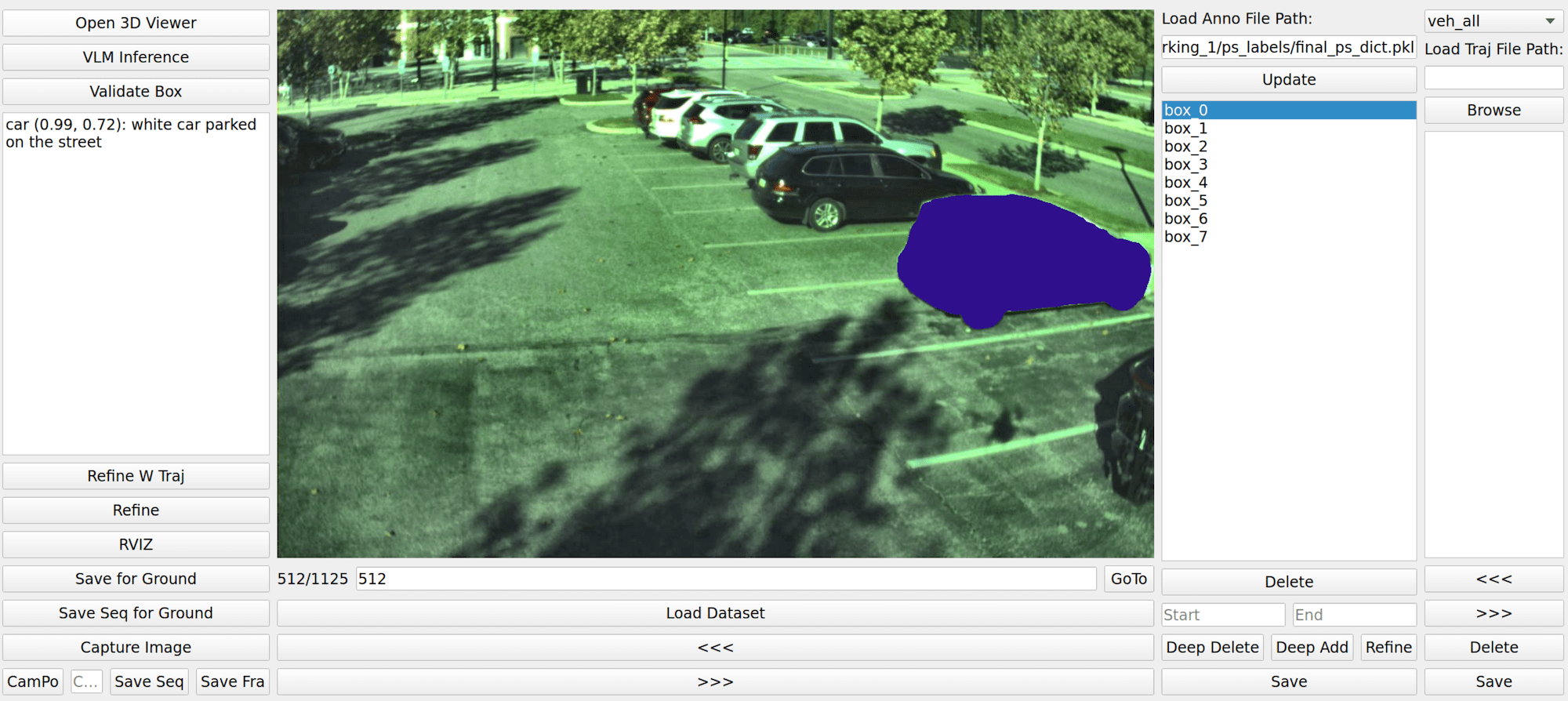}
    \vspace{-0.5cm}
    \caption{Automatic pseudo-label screening interface powered by the Tokenize Anything model.}
\label{fig:bbox_ui}
\end{figure}

\noindent\textbf{Drone and Quadruped Platforms.} For the {\includegraphics[width=0.028\linewidth]{figures/icons/drone.png}} \texttt{Drone} and {\includegraphics[width=0.022\linewidth]{figures/icons/quadruped.png}} \texttt{Quadruped} platform, the original M3ED Dataset~\cite{chaney2023m3ed} does not contain pre-annotated 3D bounding boxes and require custom 3D bounding box annotations. We establish an annotation pipeline introduced in Figure~2 of the main paper. The process is composed of three stages:

\begin{itemize}
    \item \emph{Pseudo-label seeding.} We first pretrain a diverse set of state-of-the-art 3D detectors: PV-RCNN~\cite{shi2020pv}, PV-RCNN++~\cite{shi2023pv}, Voxel-RCNN~\cite{deng2021voxel}, IA-SSD~\cite{zhang2022not}, CenterPoint~\cite{yin2021center}, and SECOND~\cite{yan2018second}, on large-scale external datasets (\eg, Waymo~\cite{Sun_2020_CVPR}, nuScenes~\cite{caesar2020nuscenes}, Lyft~\cite{houston2021one}). These models are then used to infer pseudo-labels on our data, covering a variety of sensor configurations and scene layouts.

    \item \noindent\emph{Automatic consolidation.} To consolidate predictions, we apply a kernel density estimation (KDE) approach to fuse overlapping boxes and improve consistency. A 3D multi-object tracking algorithm (CTRL~\cite{fan2023once}) is used to propagate detections over time and interpolate missing instances. To further validate category correctness, we employ the Tokenize Anything model~\cite{pan2024tokenize} to project pseudo-boxes onto RGB images and cross-check the detected objects with open-vocabulary tags (see Figure~\ref{fig:bbox_ui}). Boxes with mismatched semantics are flagged for review, reducing semantic drift across modalities.

    \item \noindent\emph{Human refinement.} Finally, we manually refine each box on a per-frame basis. Three trained annotators iteratively verify, correct, and cross-validate all annotations to ensure high-quality outputs. Despite the assistance from automation, the sparsity and noise of real-world point clouds require human oversight. 
\end{itemize}
This multi-stage toolkit integrates detection, filtering, image-level verification, and annotation interfaces. It enables scalable and accurate labeling for mobile platforms where no prior annotations exist, contributing to the high consistency and realism of our dataset.

\subsubsection{Annotation Details on Referring Expressions}

\begin{table*}[t]
    \centering
    \caption{Prompt for Single-Object Grounding}
    \vspace{-0.2cm}
    \renewcommand{\arraystretch}{1.3}
    \setlength{\tabcolsep}{10pt}
    \tcbset{
        colback=gray!10,
        colframe=gray!50,
        arc=2mm,
        boxrule=0.3mm,
        width=\textwidth,
    }
    \begin{tcolorbox}
        \textbf{You are an assistant designed to generate fine-grained descriptions for 3D objects grounded in images.} \\\\

        Given a single object highlighted by a bounding box and its class label, please generate a detailed and unambiguous description focusing on the following aspects:
        \begin{itemize}
            \item \textbf{1. Class}: Specify the object’s type and visual features (e.g., color, shape, vehicle model, clothing of pedestrians).
            \item \textbf{2. Status}: Indicate whether the object is static or in motion, and describe its speed or behavioral state.
            \item \textbf{3. Absolute Position}: Describe the object's location within the image (e.g., bottom-left, center).
            \item \textbf{4. Viewer Perspective}: Explain the object’s orientation relative to the camera or viewer (e.g., facing the camera, viewed from behind).
            \item \textbf{5. Spatial Relations}: Outline how the object is situated relative to nearby elements in the scene.
            \item \textbf{6. Moving Direction} (if applicable): Specify whether the object is moving toward or away from the viewer, or turning in a particular direction.
        \end{itemize}

        \vspace{0.2cm}
        After addressing each aspect, \textbf{compose a fluent summary sentence (less than 100 words)} that uniquely identifies the object within the scene.

        \vspace{0.2cm}
        \textbf{Response Format:}
        \begin{verbatim}
    1. class: [...]
    2. status: [...]
    3. position in the image: [...]
    4. relation to the viewer: [...]
    5. relationships with other objects: [...]
    6. moving direction: [...]
    7. Summary: [complete descriptive sentence]
        \end{verbatim}

        \vspace{0.2cm}
        \textbf{Important:} Your description should be as specific and detailed as possible. Ensure the response is uniquely aligned with the given object and avoids ambiguity.
    \end{tcolorbox}
    \label{tab:prompt_single_refined}
\end{table*}

\begin{table*}[t]
    \centering
    \caption{Prompt for Multi-Object Grounding}
    \vspace{-0.2cm}
    \renewcommand{\arraystretch}{1.3}
    \setlength{\tabcolsep}{10pt}
    \tcbset{
        colback=gray!10,
        colframe=gray!50,
        arc=2mm,
        boxrule=0.3mm,
        width=\textwidth,
    }
    \begin{tcolorbox}
        \textbf{You are a multimodal assistant tasked with describing and comparing two objects in a temporal 3D scene.} \\\\

        You are provided with a sequence of images where two objects are marked with green bounding boxes. You will also be given:
        \begin{itemize}
            \item The class label of each object
            \item A predefined semantic relationship between them
        \end{itemize}

        Your task is to describe \textbf{each object individually}, and then articulate the relationship between them. Ensure your descriptions are \textbf{precise, grounded in visual evidence}, and cover the following perspectives:
        \begin{itemize}
        \item \textbf{1. Appearance}: Describe the object’s color, texture, size (small, medium, large), shape, category, and material.
        \item \textbf{2. State}: Specify whether the object is moving or static, and describe its current action (e.g., turning, accelerating).
        \item \textbf{3. Spatial Relationship}: Explain its location and relation to nearby scene elements.
        \item \textbf{4. Temporal Movement}: Summarize how the object’s position changes across the image sequence.
        \item \textbf{5. Other}: Include any other details that can aid recognition.
        \end{itemize}

        \textbf{Then, describe the relationship between the two objects} based on their relative spatial or temporal behavior (e.g., “the car is overtaking the cyclist”, “the robot is approaching the chair”).

        \vspace{0.2cm}
        \textbf{Response Format:}
        \begin{verbatim}
    Object A:
    1. appearance: [...]
    2. state: [...]
    3. spatial relationship: [...]
    4. temporal movement: [...]
    5. other: [...]
    
    Object B:
    1. appearance: [...]
    2. state: [...]
    3. spatial relationship: [...]
    4. temporal movement: [...]
    5. other: [...]
    
    Relationship: [description of how Object A relates to Object B]
        \end{verbatim}

        \textbf{Important:} Focus only on the two marked objects. Your response must be detailed and unambiguous, and should accurately reflect both visual and temporal information.
    \end{tcolorbox}
    \label{tab:prompt_multi_refined}
\end{table*}

To evaluate grounding performance under natural and unambiguous language, we annotate referring expressions for each 3D bounding box in our dataset. These expressions are designed to support both single-object and multi-object grounding across diverse platforms, and are generated via a hybrid automatic–manual pipeline.

\noindent\textbf{Generation with Vision-Language Models.}  
We use the Qwen2-VL-72B~\cite{qwen2-vl} vision-language model to automatically generate initial referring expressions. For each annotated 3D bounding box, we first project it onto the corresponding RGB image frame, then provide both the image and a task-specific prompt to the model. The prompts are carefully designed to guide the model to produce detailed, visually grounded, and unambiguous expressions.

For the \emph{single-object grounding} setting, we use a structured prompt (see Table~\ref{tab:prompt_single_refined}) that elicits descriptions covering the object's class, status, absolute position, spatial relationships, and motion. For the \emph{multi-object grounding} setting, we adopt a more compositional prompt (see Table~\ref{tab:prompt_multi_refined}) that encourages descriptions of two objects and their semantic relationships in temporal 3D scenes, covering appearance, motion, and relative spatial configuration.

\begin{figure}[t]
    \centering
    \includegraphics[width=\linewidth]{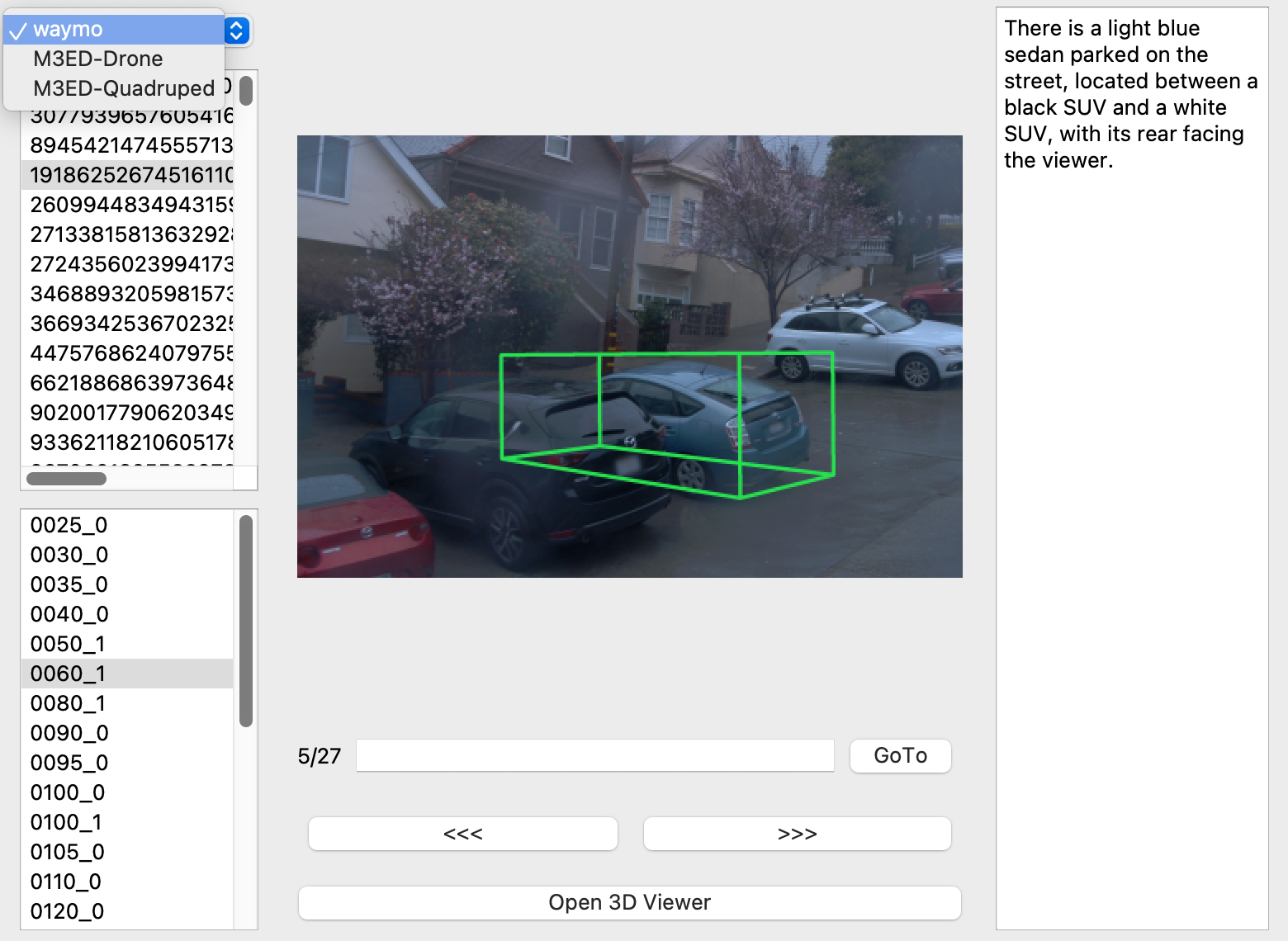}
    \vspace{-0.5cm}
    \caption{Graphical user interface used during the human refinement phase. Annotators inspect each scene by viewing the 3D bounding box projected onto the RGB image, alongside the automatically generated referring expression. Annotators verify or revise the description to ensure it uniquely and accurately identifies the target object. Scenes failing this verification are discarded.}
\label{fig:human_refinement_ui}
\end{figure}

\noindent\textbf{Manual Verification and Filtering.}  
All generated referring expressions undergo human verification to ensure semantic correctness, referential clarity, and linguistic fluency. To facilitate this process, we develop a custom annotation interface, as shown in Figure~\ref{fig:human_refinement_ui}. Annotators review each expression in the context of the full scene, with the target object visualized via its projected 3D bounding box overlaid on the RGB image. If an expression is partially inaccurate or omits essential details, it may be directly edited. If the description is fundamentally flawed -- such as containing hallucinated attributes or being referentially ambiguous -- the sample is discarded. This verification process is conducted by a team of five trained annotators to ensure consistency and overall annotation quality.

\noindent\textbf{Platform-Aware Annotation Alignment.}  
To support fair and consistent evaluation across diverse platforms, we adopt a unified annotation protocol for {\includegraphics[width=0.021\linewidth]{figures/icons/vehicle.png}} \texttt{Vehicle}, {\includegraphics[width=0.028\linewidth]{figures/icons/drone.png}} \texttt{Drone}, and {\includegraphics[width=0.022\linewidth]{figures/icons/quadruped.png}} \texttt{Quadruped} scenes. Specifically, the same instruction prompt is used across all platforms, ensuring that the generation process follows identical linguistic and visual grounding expectations, regardless of the underlying sensor configuration or viewpoint.

All spatial descriptions in referring expressions are written from the \emph{observer’s perspective}, \ie, relative to the camera view that captured the scene. This design allows language like ``on the left'', ``facing away'', or ``in the front'' to remain intuitive and unambiguous to models operating on image-grounded or LiDAR-centered input. Rather than using global scene-relative coordinates (\eg, ``north-east corner''), we ensure all position statements are grounded in the visual evidence available from the sensor’s viewpoint.

\begin{figure}[t]
    \centering
    \includegraphics[width=\linewidth]{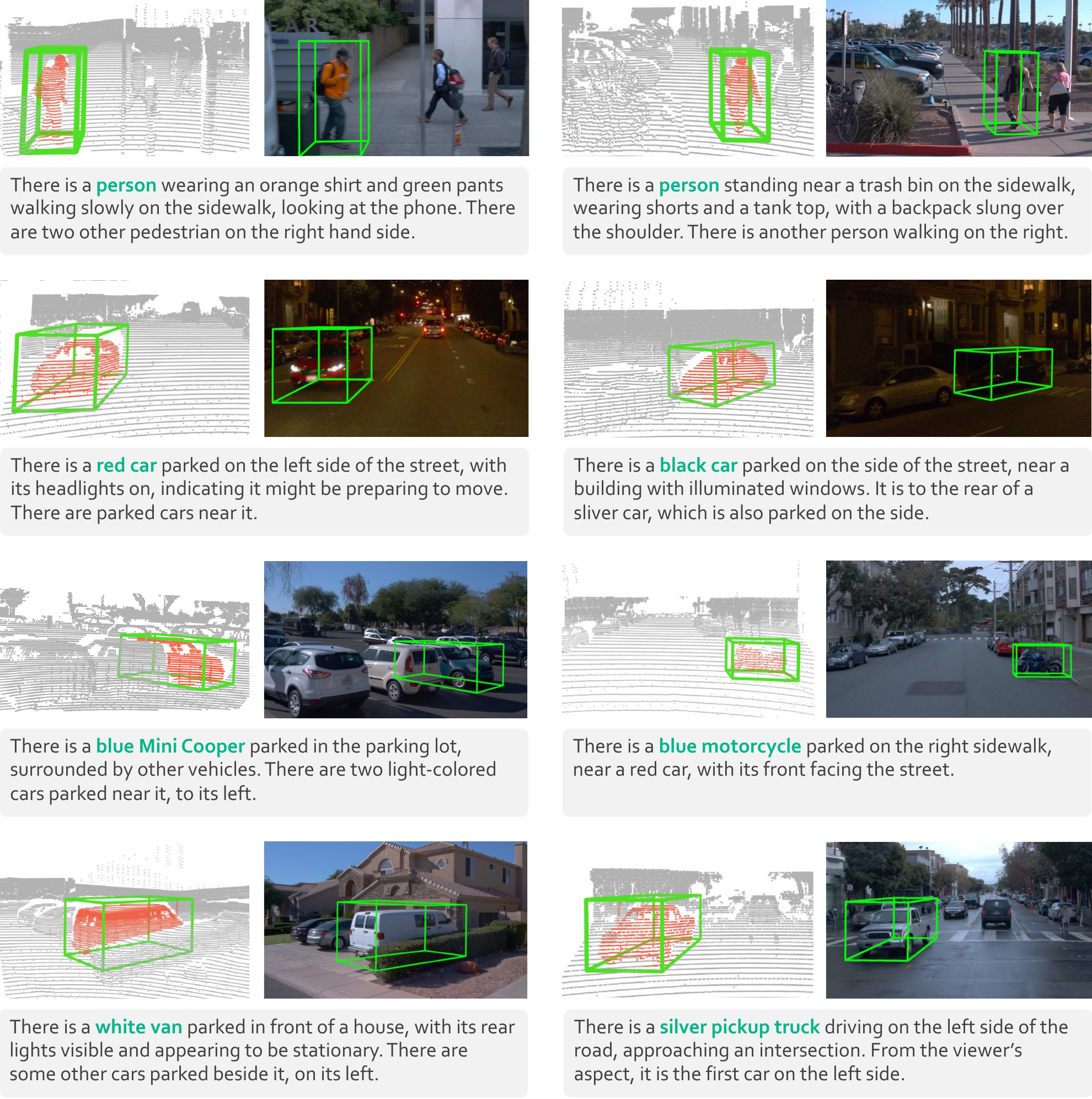}
    \vspace{-0.4cm}
    \caption{\textbf{Additional examples of 3D grounding} from the {\includegraphics[width=0.021\linewidth]{figures/icons/vehicle.png}} \texttt{Vehicle} platform in \ours~dataset. The data shown include the LiDAR point clouds, the RGB frames, and the associated referring expressions. Best viewed in colors and zoomed in for more details.}
\label{fig:supp_examples_single_vehicle}
\end{figure}

\begin{figure}[t]
    \centering
    \includegraphics[width=\linewidth]{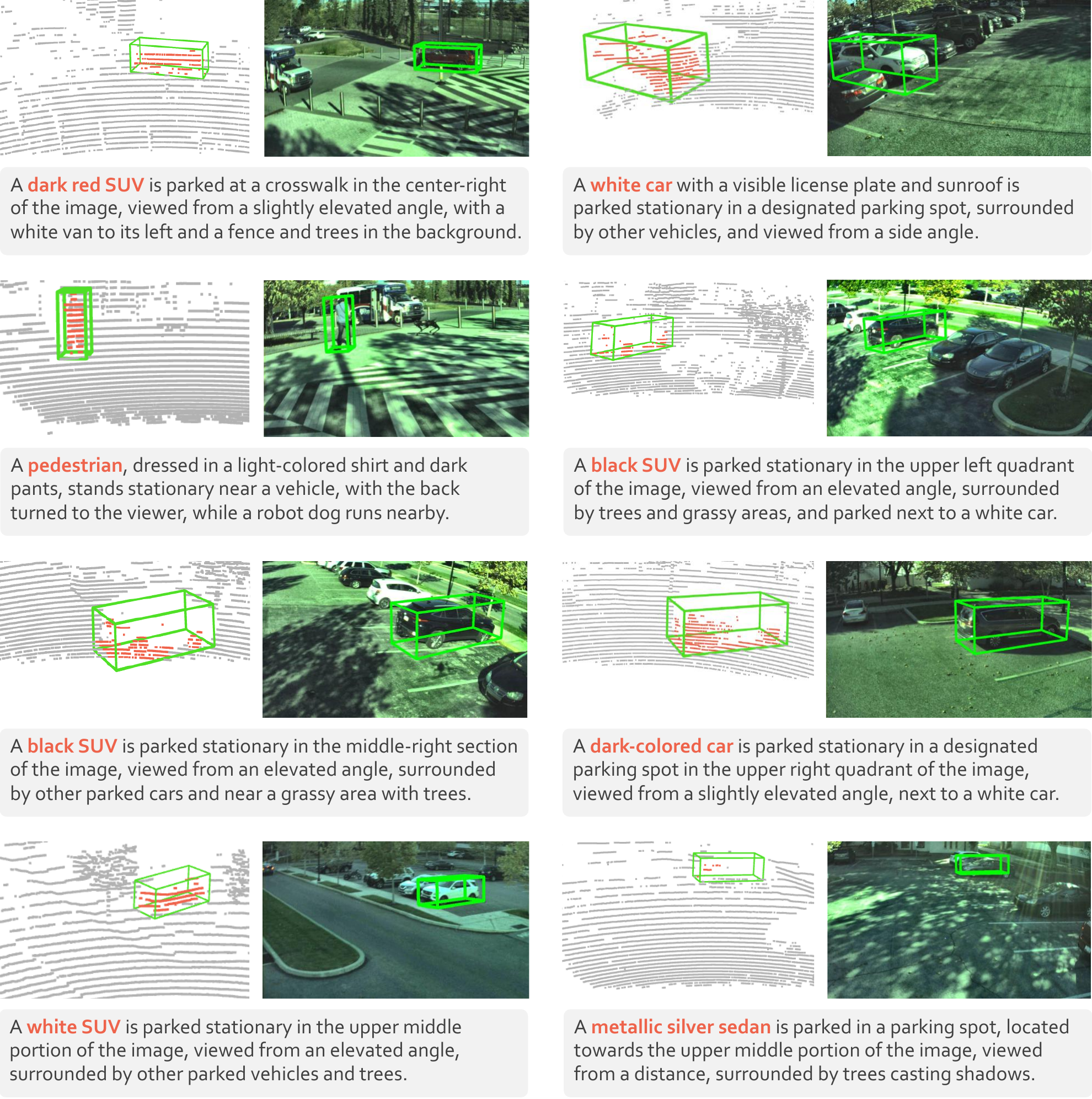}
    \vspace{-0.4cm}
    \caption{\textbf{Additional examples of 3D grounding} from the {\includegraphics[width=0.028\linewidth]{figures/icons/drone.png}} \texttt{Drone} platform in \ours~dataset. The data shown include the LiDAR point clouds, the RGB frames, and the associated referring expressions. Best viewed in colors and zoomed in for more details.}
\label{fig:supp_examples_single_drone}
\end{figure}

\begin{figure}[t]
    \centering
    \includegraphics[width=\linewidth]{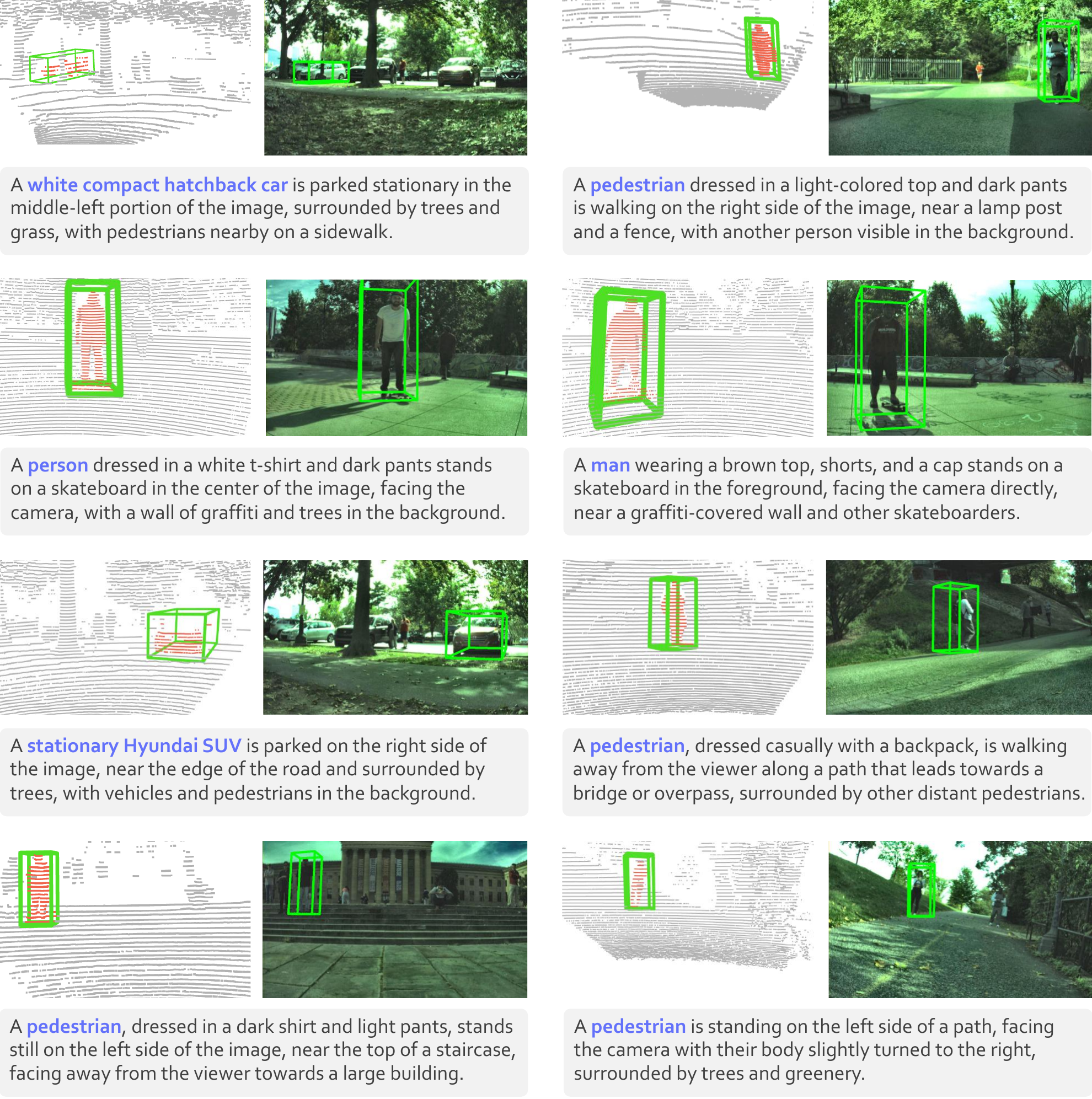}
    \vspace{-0.4cm}
    \caption{\textbf{Additional examples of 3D grounding} from the {\includegraphics[width=0.022\linewidth]{figures/icons/quadruped.png}} \texttt{Quadruped} platform in \ours~dataset. The data shown include the LiDAR point clouds, the RGB frames, and the associated referring expressions. Best viewed in colors and zoomed in for more details.}
\label{fig:supp_examples_single_quadruped}
\end{figure}

\begin{figure}[t]
    \centering
    \includegraphics[width=\linewidth]{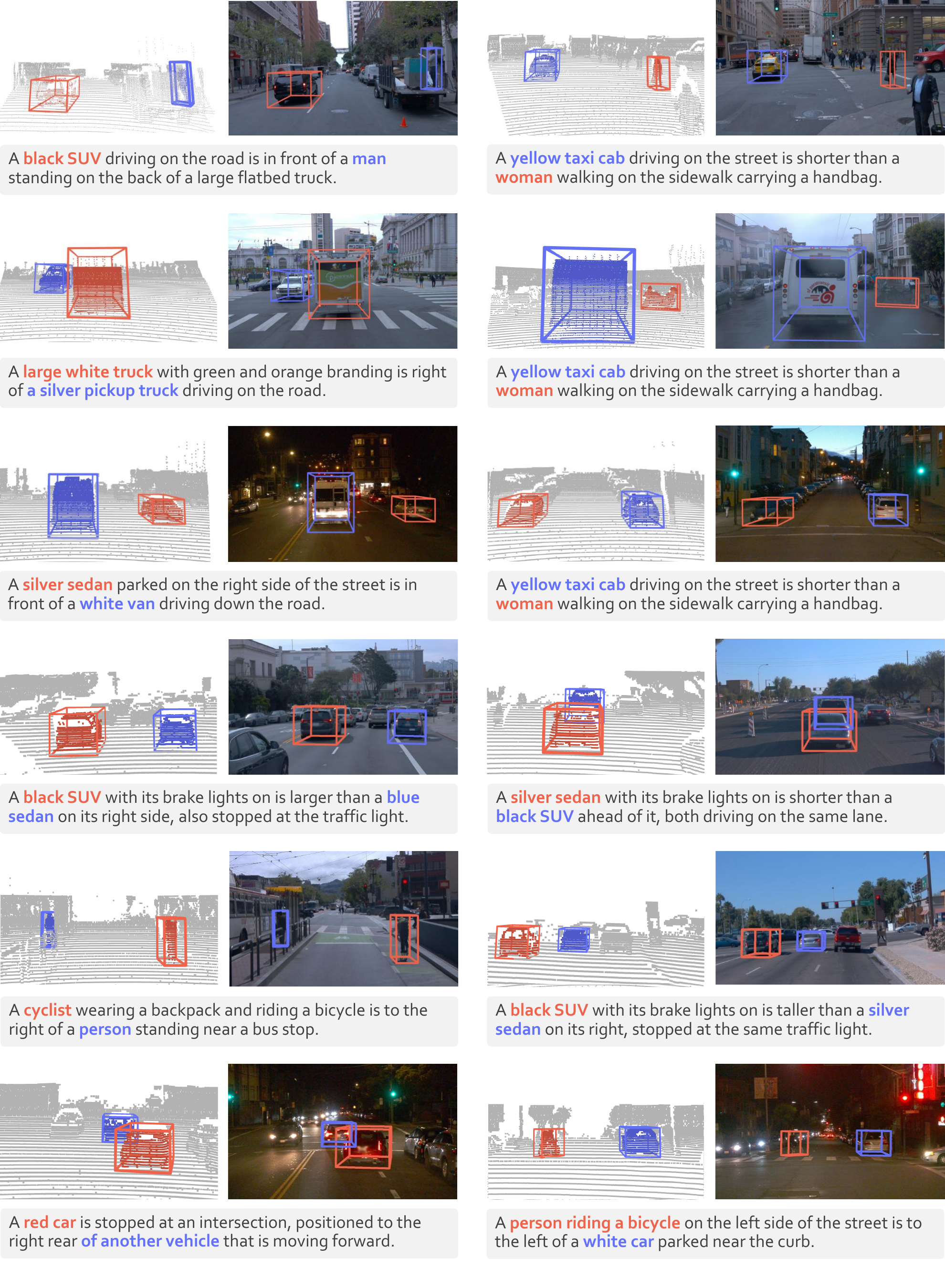}
    \vspace{-0.4cm}
    \caption{\textbf{Additional examples of multi-object 3D grounding} from the \ours~dataset. }
\label{fig:supp_examples_multi}
\end{figure}

\subsection{Examples of Single-Object 3D Grounding}

Figure~\ref{fig:supp_examples_single_vehicle}, Figure~\ref{fig:supp_examples_single_drone}, and Figure~\ref{fig:supp_examples_single_quadruped} present representative examples of single-object 3D grounding from the {\includegraphics[width=0.021\linewidth]{figures/icons/vehicle.png}} \texttt{Vehicle}, {\includegraphics[width=0.028\linewidth]{figures/icons/drone.png}} \texttt{Drone}, and {\includegraphics[width=0.022\linewidth]{figures/icons/quadruped.png}} \texttt{Quadruped} platforms in our dataset. Each example displays the fused RGB image and LiDAR point cloud, along with a natural language referring expression and its corresponding 3D bounding box.

These examples highlight several key characteristics of the \ours~dataset:

\begin{itemize}
    \item \emph{Cross-platform diversity.} {\includegraphics[width=0.021\linewidth]{figures/icons/vehicle.png}} \texttt{Vehicle} scenes often feature structured road layouts with multiple traffic participants, such as cars, pedestrians, and motorcycles. {\includegraphics[width=0.028\linewidth]{figures/icons/drone.png}} \texttt{Drone} scenes offer wide-area top-down coverage with more cluttered object distributions, including overlapping vehicles, elevated viewpoints, and richer spatial context. {\includegraphics[width=0.022\linewidth]{figures/icons/quadruped.png}} \texttt{Quadruped} scenes are recorded from a low-altitude, ground-level perspective, focusing on close-range human interactions and sidewalk-level details.
    
    \item \emph{Natural language variation.} Referring expressions reflect platform-specific visibility and spatial reasoning. For example, {\includegraphics[width=0.021\linewidth]{figures/icons/vehicle.png}} \texttt{Vehicle} -mounted viewpoints encourage descriptions like ``on the left side of the street'', while {\includegraphics[width=0.028\linewidth]{figures/icons/drone.png}} \texttt{Drone}-based annotations describe objects ``in the upper right quadrant'' or ``viewed from above''. {\includegraphics[width=0.022\linewidth]{figures/icons/quadruped.png}} \texttt{Quadruped} expressions capture nuanced positional cues (\eg, ``facing the camera'', ``walking away on the path'') and often describe subtle behaviors or clothing.

    \item \emph{Scene conditions.} Our dataset includes scenes captured under diverse environmental conditions, including both daytime and nighttime settings. This is evident in the {\includegraphics[width=0.021\linewidth]{figures/icons/vehicle.png}} \texttt{Vehicle} and {\includegraphics[width=0.028\linewidth]{figures/icons/drone.png}} \texttt{Drone} examples, where objects may be illuminated by streetlights or appear in low-light settings, adding realism and complexity to the grounding task.

    \item \emph{Multi-modal alignments.} Despite differences in viewpoint and density, all annotations maintain strong visual-language grounding. Each expression unambiguously describes a target object with sufficient detail for model disambiguation, including appearance, position, context, and motion when applicable.
\end{itemize}

These examples demonstrate the richness and difficulty of grounding in our dataset: models must generalize across platforms, lighting conditions, and spatial perspectives while maintaining consistent language understanding. The platform-aware yet prompt-consistent annotation pipeline ensures comparability while preserving diversity.

\subsection{Examples of Multi-Object 3D Grounding}

Figure~\ref{fig:supp_examples_multi} presents representative examples from the multi-object grounding subset of our dataset. In this setting, each scene contains two target objects annotated with distinct 3D bounding boxes and described through interrelated referring expressions. These expressions not only characterize each object individually (\eg, class, appearance, motion), but also explicitly capture their spatial, temporal, or semantic relationships.

The examples span a variety of real-world outdoor scenarios involving pedestrians, cyclists, and vehicles. Referring expressions encode rich visual-semantic grounding cues, such as:
\begin{itemize}
  \item \textbf{Relative positioning}: ``in front of'', ``to the right of'', ``ahead of'', ``shorter than''.
  \item \textbf{Comparative reasoning}: ``is larger than'', ``is taller than'', ``is shorter than''.
  \item \textbf{Temporal context and motion state}: ``driving on the road'', ``stopped at the traffic light'', ``moving forward''.
\end{itemize}

\subsection{Statistics and Analyses}

In this section, we present detailed statistics and analyses that characterize the \ours~dataset across platforms and splits. We examine the distribution of scene complexity, defined by the number of annotated objects per scene, and show how this varies significantly between the {\includegraphics[width=0.021\linewidth]{figures/icons/vehicle.png}} \texttt{Vehicle}, {\includegraphics[width=0.028\linewidth]{figures/icons/drone.png}} \texttt{Drone}, and {\includegraphics[width=0.022\linewidth]{figures/icons/quadruped.png}} \texttt{Quadruped} platforms. 
Additionally, we analyze point-level density within 3D bounding boxes, highlighting strong differences in LiDAR sampling resolution across platforms. These statistics provide important context for interpreting grounding performance and understanding platform-specific challenges in 3D perception and language grounding.

\subsubsection{Scene Complexity Statistics across Platforms}

Table~\ref{tab:scene_count_combined} presents detailed statistics of the training and validation splits across the three platforms in the \ours~dataset -- {\includegraphics[width=0.021\linewidth]{figures/icons/vehicle.png}} \texttt{Vehicle}, {\includegraphics[width=0.028\linewidth]{figures/icons/drone.png}} \texttt{Drone}, {\includegraphics[width=0.022\linewidth]{figures/icons/quadruped.png}} \texttt{Quadruped} platforms. Each scene is categorized by the number of objects it contains, providing insight into the distribution of scene complexity. These statistics are collected on the single-object grounding subset, where only one referred object is annotated per scene.

We observe that {\includegraphics[width=0.022\linewidth]{figures/icons/quadruped.png}} \texttt{Quadruped} scenes are predominantly low-density: in the combined training+validation split, $59.7$\% of scenes contain fewer than seven targets ($1$–$3$: $31.4$\%, $4$–$6$: $28.5$\%). Such low-density settings simplify the localization task and reduce ambiguity during reference resolution. In contrast, {\includegraphics[width=0.028\linewidth]{figures/icons/drone.png}} \texttt{Drone} data features a much higher proportion of crowded scenes: about $50\%$ of the training scenes and $62\%$ of the validation scenes contain $7$ or more objects. This reflects the broader aerial perspective and wider field of view, which captures more complex environments and increases grounding difficulty.

The {\includegraphics[width=0.021\linewidth]{figures/icons/vehicle.png}} \texttt{Vehicle} platform lies between the two, exhibiting a relatively balanced distribution of scene complexities. This makes Vehicle data a valuable middle ground for learning models that generalize across both sparse and dense settings.

Overall, these statistics highlight the diverse spatial configurations in our dataset and provide context for the performance variations discussed in the experiment section of the main paper, particularly in the cross-platform grounding evaluation.

\begin{table}[t]
\centering
\caption{Scene count grouped by number of objects per scene across platforms and splits.}
\label{tab:scene_count_combined}
\vspace{0.1cm}
\renewcommand{\arraystretch}{1.2}
\setlength{\tabcolsep}{12pt}
\begin{tabular}{l|ccccc|c}
\toprule
\textbf{Platform} & \textbf{1--3} & \textbf{4--6} & \textbf{7--9} & \textbf{10--12} & \textbf{13+} & \textbf{Total} \\
\midrule\midrule
\rowcolor{3eed_green!15}\multicolumn{7}{l}{\textcolor{3eed_green}{\textit{\textbf{Training}}}}  \\
\quad {\includegraphics[width=0.021\linewidth]{figures/icons/vehicle.png}} \texttt{Vehicle}     & $1{,}177$ & $968$ & $360$ & $135$ & $61$ & $2{,}701$ \\
\quad {\includegraphics[width=0.028\linewidth]{figures/icons/drone.png}} \texttt{Drone}         & $1{,}021$ & $1{,}053$ & $1{,}035$ & $265$ & $740$ & $4{,}114$ \\
\quad {\includegraphics[width=0.022\linewidth]{figures/icons/quadruped.png}} \texttt{Quadruped} & $1{,}614$ & $1{,}528$ & $1{,}263$ & $527$ & $0$ & $4{,}932$ \\
\quad \textbf{Total} & $\mathbf{3{,}812}$ & $\mathbf{3{,}549}$ & $\mathbf{2{,}658}$ & $\mathbf{927}$ & $\mathbf{801}$ & $\mathbf{11{,}747}$ \\
\midrule
\rowcolor{3eed_green!15}\multicolumn{7}{l}{\textcolor{3eed_green}{\textit{\textbf{Validation}}}}\\
\quad {\includegraphics[width=0.021\linewidth]{figures/icons/vehicle.png}} \texttt{Vehicle}     & $1{,}154$ & $927$ & $403$ & $180$ & $44$ & $2{,}708$ \\
\quad {\includegraphics[width=0.028\linewidth]{figures/icons/drone.png}} \texttt{Drone}         & $411$ & $734$ & $494$ & $599$ & $746$ & $2{,}984$ \\
\quad {\includegraphics[width=0.022\linewidth]{figures/icons/quadruped.png}} \texttt{Quadruped} & $855$ & $696$ & $530$ & $837$ & $10$ & $2{,}928$ \\
\quad \textbf{Total} & $\mathbf{2{,}420}$ & $\mathbf{2{,}357}$ & $\mathbf{1{,}427}$ & $\mathbf{1{,}616}$ & $\mathbf{800}$ & $\mathbf{8{,}620}$ \\
\midrule
\rowcolor{darkgray!15}\textbf{Summary} & $\mathbf{6{,}232}$ & $\mathbf{5{,}906}$ & $\mathbf{4{,}085}$ & $\mathbf{2{,}543}$ & $\mathbf{1{,}601}$ & $\mathbf{20{,}367}$ \\
\bottomrule
\end{tabular}
\end{table}

\subsubsection{Box Density Statistics}

\begin{figure}[t]
    \centering
    \includegraphics[width=\linewidth]{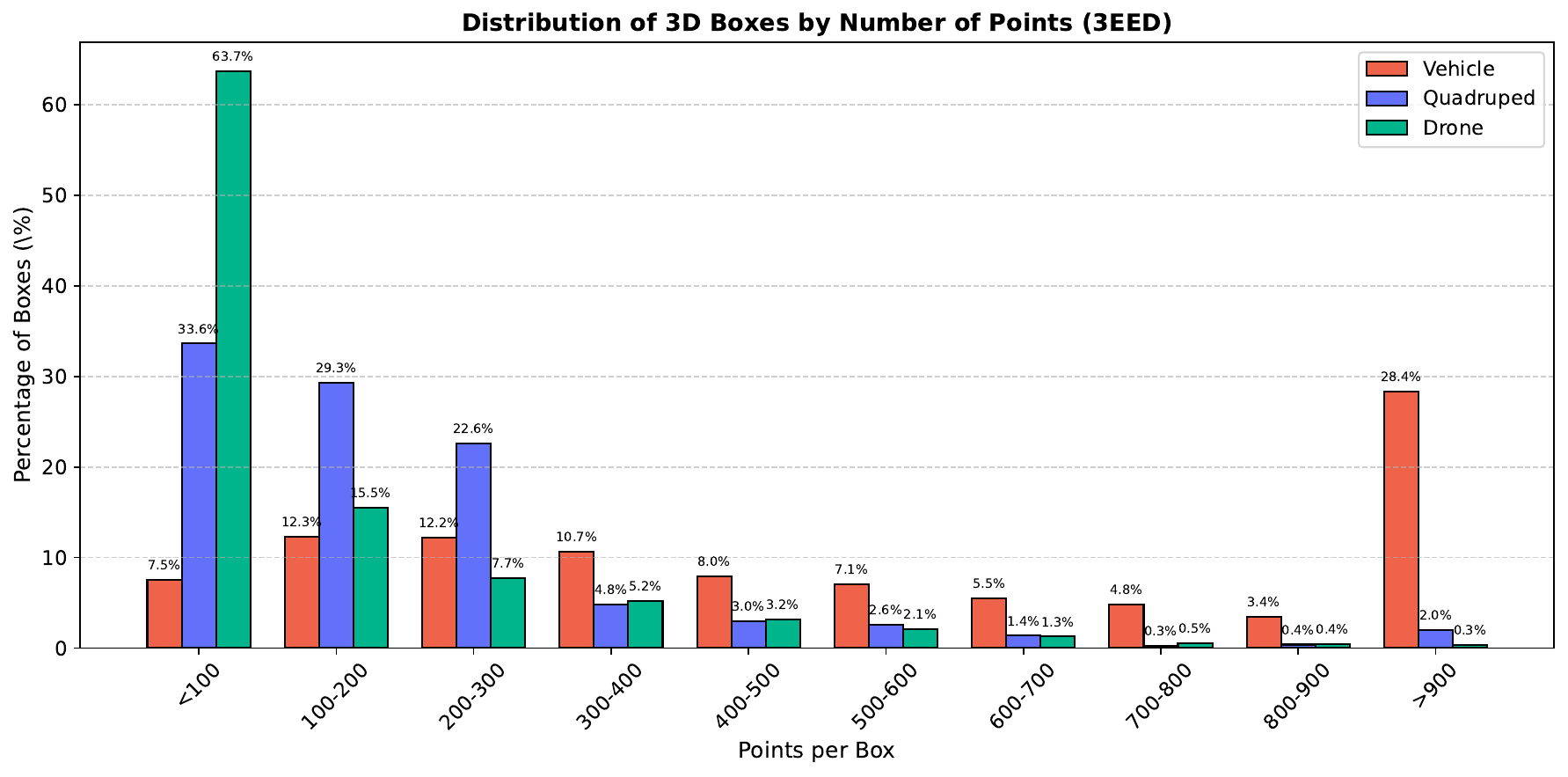}
    \vspace{-0.5cm}
    \caption{\textbf{Distribution of 3D boxes by number of points contained in each box}, across {\includegraphics[width=0.021\linewidth]{figures/icons/vehicle.png}} \texttt{Vehicle}, {\includegraphics[width=0.028\linewidth]{figures/icons/drone.png}} \texttt{Drone}, and {\includegraphics[width=0.022\linewidth]{figures/icons/quadruped.png}} \texttt{Quadruped} platforms. Drone boxes are significantly sparser, while Vehicle boxes are generally denser, indicating strong variations in point cloud density across platforms.}
    \label{fig:points_per_box_distribution}
\end{figure}

Figure~\ref{fig:points_per_box_distribution} illustrates the distribution of 3D bounding boxes by the number of LiDAR points contained within each box, across the three platforms. The {\includegraphics[width=0.028\linewidth]{figures/icons/drone.png}} \texttt{Drone} platform features extremely sparse boxes, with over $60\%$ containing fewer than $100$ points. This is a result of its high-altitude viewpoint and long-range perception, which leads to sparser spatial sampling. Conversely, the {\includegraphics[width=0.021\linewidth]{figures/icons/vehicle.png}} \texttt{Vehicle}  platform has more than $28\%$ of boxes with over $900$ points, reflecting the dense coverage typical in street-level LiDAR. The {\includegraphics[width=0.022\linewidth]{figures/icons/quadruped.png}} \texttt{Quadruped} platform occupies a middle ground but still exhibits noticeable sparsity, with a third of its boxes containing fewer than $100$ points.

These density differences strongly affect 3D feature quality and grounding performance, especially in low-point regimes where accurate object localization becomes more challenging. 

\subsection{License}
The \ours~dataset and its associated toolkit are released under the Attribution-ShareAlike 4.0 International (CC BY-SA 4.0)\footnote{\url{https://creativecommons.org/licenses/by-sa/4.0/legalcode}.} license.

\section{Benchmark Construction Details}

In this section, we describe how we construct benchmark settings for evaluating 3D language grounding using our dataset.
All tasks are formulated in a proposal-free setting, where models must directly predict 3D bounding boxes from point clouds and referring expressions. 
We also detail the baseline models, training configurations, and evaluation metrics used throughout our experiments.
Our goal is to enable fair, controlled, and reproducible comparison across grounding tasks with varying spatial and linguistic complexity.

\subsection{Single-Object Grounding Baselines}

We compare our approach against two 3D visual grounding baselines adapted to the outdoor point cloud domain: \textbf{BUTD-DETR}~\cite{jain2022butd-detr} and \textbf{EDA}~\cite{wu2023eda}. Both models were originally proposed for grounding in 3D indoor scenes~\cite{dai2017scannet}, and we adapt them to our benchmark with raw point cloud input. In all comparisons, we follow a unified setting that does not rely on pre-computed object proposals; each model directly predicts 3D bounding boxes from the raw point cloud and query language.

\noindent\textbf{BUTD-DETR} \cite{jain2022butd-detr} is a transformer-based grounding model that fuses top-down language cues and bottom-up visual features for referential localization. In our setting, we remove the use of region proposals entirely and adapt the model to operate on raw point clouds. The point cloud is encoded using a PointNet++ backbone~\cite{qi2017pointnet++}, producing a sequence of 3D-aware visual tokens. The language input is processed by a frozen RoBERTa-base encoder~\cite{liu2019roberta}, generating contextualized word embeddings. The encoder module uses separate self-attention and cross-attention layers to jointly process language and visual streams. The decoder is composed of transformer layers, where non-parametric queries are derived from the top-$K$ visual tokens based on confidence scores. Each query outputs a 3D bounding box via a regression head that predicts box center and size relative to the anchor point. It supervises the model using a Hungarian matching algorithm that assigns queries to ground-truth boxes. 
We retain the original box regression and token-level soft alignment loss. The contrastive loss is also included, with a symmetric formulation that aligns all predicted queries to token embeddings and vice versa, following their \emph{not-mentioned} augmentation strategy for unmatched queries.

\noindent\textbf{EDA} \cite{wu2023eda} decomposes each language query into semantic components and explicitly aligns them with point-level features. The model uses the same point encoder as BUTD-DETR \cite{jain2022butd-detr}. The language input is encoded via a frozen RoBERTa-base model and parsed into three components: object type, visual attributes, and spatial relations. Each component attends to the point features via separate alignment branches, predicting soft attention masks over the point cloud. The decoder aggregates these aligned components through cross-attention and predicts the final 3D bounding box via a regression head. The model is trained with a combination of L1 and GIoU losses for box prediction, along with a multi-branch semantic alignment loss that supervises the consistency between each language component and its corresponding spatial region.

\subsection{Multi-Object Grounding Baselines}
We extend the single-object grounding paradigm to handle multiple objects. Given a natural language utterance and a 3D scene, the model aims to localize all target objects referred to in the input. The core challenge lies in resolving the correspondence between multiple referred entities and their textual descriptions within the utterance.

To address this, we construct a token-level association map that aligns each target object to its corresponding span in the language input. Each object is linked to a binary mask over the token sequence, indicating which words describe it. These masks are normalized to ensure balanced supervision across all objects during training.

Hungarian matching is used to assign predictions to ground-truth boxes. In the single-object case, each scene involves a single reference box. In the multi-object case, matching is performed for each target object separately, with losses computed and averaged across targets.

During inference, the model processes a single utterance that refers to multiple target objects. For each object, we compute the semantic similarity between the candidate boxes and the relevant language span, and select top-ranked boxes based on these similarity scores.

\subsection{Implementation Details}

\noindent \textbf{Encoder-Decoder.} Our model processes raw LiDAR point clouds, which are uniformly downsampled to $16,384$ points per scene. The point cloud is encoded using a four-layer point-based encoder with multi-scale sampling (MSS) and semantic-aware fusion (SAF) modules. The model is trained from scratch without any pretraining. The radius settings for MSS are \texttt{[[$0.2, 0.8$], [$0.8, 1.6$], [$1.6, 3.2$], [$1.6, 4.8$]]}.
Text features are extracted using a frozen RoBERTa-base~\cite{kong2023robo3d} model, and projected to a 288-dimensional space via a linear projection layer to match the point cloud feature dimension. Language and visual tokens interact through three layers of bidirectional cross-attention.
A total of $1,024$ keypoints are sampled from the output of the cross-attention encoder and used as input queries to the decoder. The decoder consists of six transformer layers that iteratively refine 3D box predictions. All boxes are predicted directly from point cloud and language input.

\noindent\textbf{Loss Function.} During training, predictions are matched to ground-truth boxes via Hungarian matching as DETR~\cite{carion2020detr}, using a cost that combines box $\ell_1$ distance, 3D generalized IoU~\cite{rezatofighi2019giou}, and a soft token-level classification score. The model is supervised using a combination of classification loss, box regression loss, GIoU loss, and a contrastive alignment loss. The contrastive loss is computed between projected visual queries and language tokens using temperature-scaled cosine similarity, with supervision applied in both query-to-token and token-to-query directions. All losses are applied at the decoder outputs.

\noindent\textbf{Training Details.} We use AdamW for optimization. For single-object grounding, the learning rate is set to $1 \times 10^{-3}$ for the point encoder and $1 \times 10^{-4}$ for all other modules. Training is conducted for $100$ epochs on two NVIDIA RTX $4090$ GPUs ($24$ GB each), with a batch size of $12$ per GPU. For multi-object grounding, the learning rate is set to $1 \times 10^{-4}$ for all modules. Training is conducted for $200$ epochs on a single RTX $4090$ GPU, also with a batch size of $12$.

\subsection{Evaluation Metrics}
To assess grounding performance, we adopt standard IoU-based metrics including \textbf{Acc@$\delta$} and \textbf{mean IoU (mIoU)}.

\noindent\textbf{Accuracy@IoU$\delta$.}
Following prior works~\cite{chen2020scanrefer, achlioptas2020referit3d}, we compute the percentage of predicted 3D bounding boxes whose Intersection over Union (IoU) with the ground-truth box exceeds a threshold $\delta \in \{0.25, 0.50\}$:
\[
\text{Acc@}\delta = \frac{1}{N} \sum_{i=1}^N \mathbbm{1} \left[\text{IoU}(\hat{b}_i, b_i^{\text{gt}}) > \delta\right],
\]
where $N$ is the number of queries, $\hat{b}_i$ is the predicted box, and $b_i^{\text{gt}}$ is the ground truth.

\noindent\textbf{Mean IoU (mIoU).}
To provide a finer-grained measure of localization quality, we also report the mean IoU between the predicted and ground-truth boxes across all queries:
\[
\text{mIoU} = \frac{1}{N} \sum_{i=1}^N \text{IoU}(\hat{b}_i, b_i^{\text{gt}})~.
\]
Unlike Acc@$\delta$, which thresholds the overlap, mIoU captures continuous localization precision and is sensitive to small alignment errors.
Together, these metrics provide a comprehensive view of grounding performance under both strict and relaxed criteria.

\subsection{Evaluation Protocol}

To ensure fair and reproducible comparison across models, we standardize the evaluation protocol across four benchmark settings.
\begin{itemize}
    \item \emph{Single-platform, single-object grounding.} Models are trained and evaluated on the same platform ({\includegraphics[width=0.021\linewidth]{figures/icons/vehicle.png}} \texttt{Vehicle}, {\includegraphics[width=0.028\linewidth]{figures/icons/drone.png}} \texttt{Drone}, and {\includegraphics[width=0.022\linewidth]{figures/icons/quadruped.png}} \texttt{Quadruped}), enabling assessment of in-domain performance under consistent sensor geometry and point cloud density. A prediction is considered correct if the predicted bounding box has an Intersection over Union (IoU) above a predefined threshold with the ground-truth box.

    \item \emph{Cross-platform transfer.} In this setting, models are trained on one platform and evaluated on a disjoint target platform (e.g., train on {\includegraphics[width=0.021\linewidth]{figures/icons/vehicle.png}} \texttt{Vehicle}, test on {\includegraphics[width=0.028\linewidth]{figures/icons/drone.png}} \texttt{Drone}). The evaluation protocol mirrors that of the single-object setting, enabling controlled assessment of cross-platform generalization.

    \item \emph{Multi-object grounding.} For queries referring to multiple objects within a scene, the model must predict all corresponding 3D bounding boxes. A prediction is deemed correct only if \emph{all} referred objects are correctly localized with IoU above the threshold. This setting tests the model’s ability to handle complex referential expressions and object-object relationships.

    \item \emph{Multi-platform grounding.} Models are trained jointly on data from all three platforms and evaluated separately on each one. This setting examines the model's robustness to diverse spatial distributions, sensor configurations, and environmental conditions in a unified training regime.
\end{itemize}

\paragraph{Reproducibility.} All evaluations are conducted on a fixed validation split with no overlap between training and evaluation scenes. The evaluation pipeline is standardized across all settings, and we release our full codebase and configuration files to support reproducible benchmarking and future comparisons.

\begin{figure}[t]
    \centering
    \includegraphics[width=\linewidth]{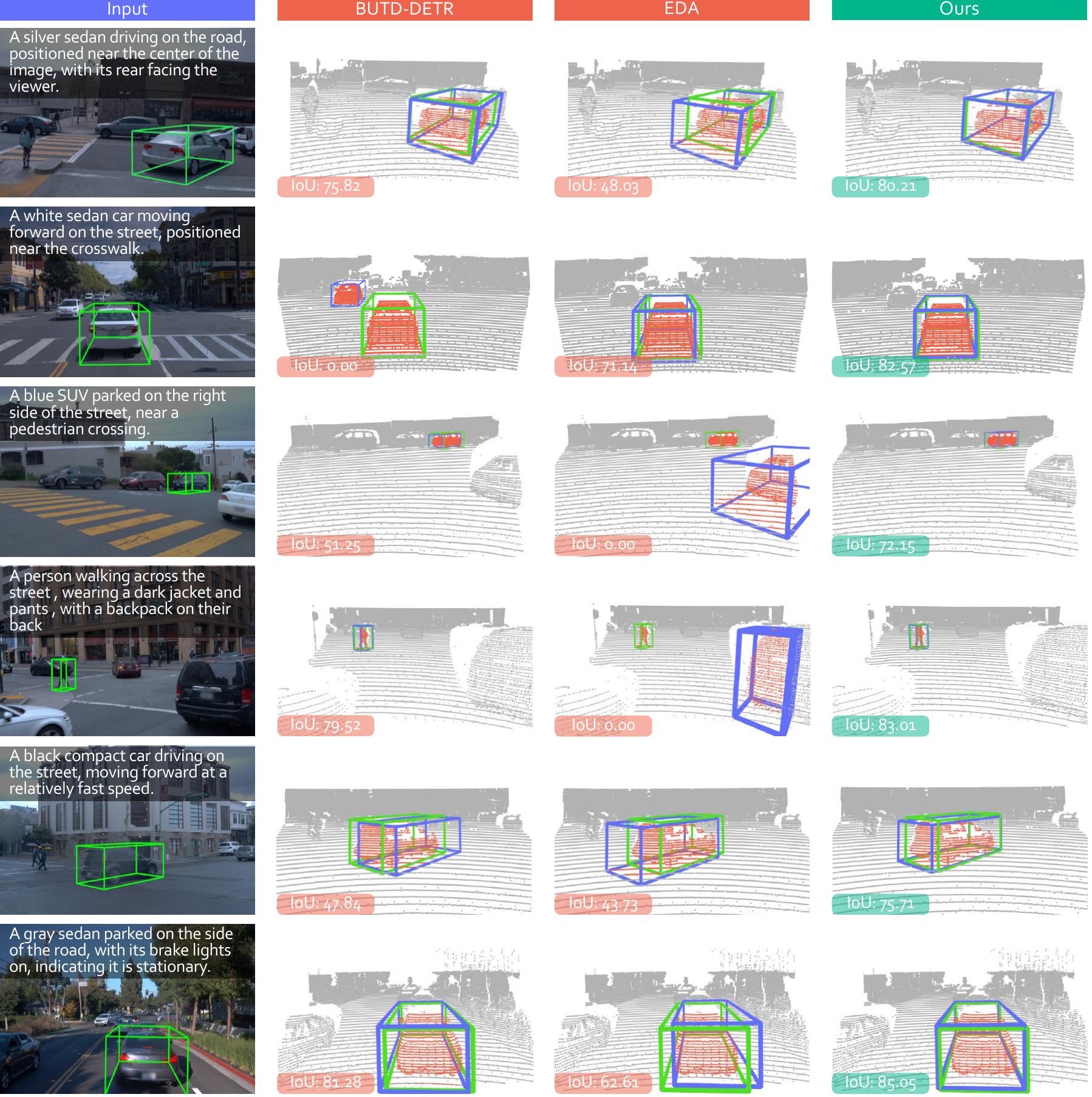}
    \vspace{-0.4cm}
    \caption{\textbf{Additional qualitative comparisons} of single-object 3D grounding on the {\includegraphics[width=0.021\linewidth]{figures/icons/vehicle.png}} \texttt{Vehicle} platform from the \ours~dataset. The data shown include the RGB frames, the LiDAR point clouds, and the associated referring expressions. The ground truth and predicted boxes are shown in green and blue, respectively. Best viewed in colors and zoomed in for more details.}
\label{fig:supp_single_vehicle}
\end{figure}

\begin{figure}[t]
    \centering
    \includegraphics[width=\linewidth]{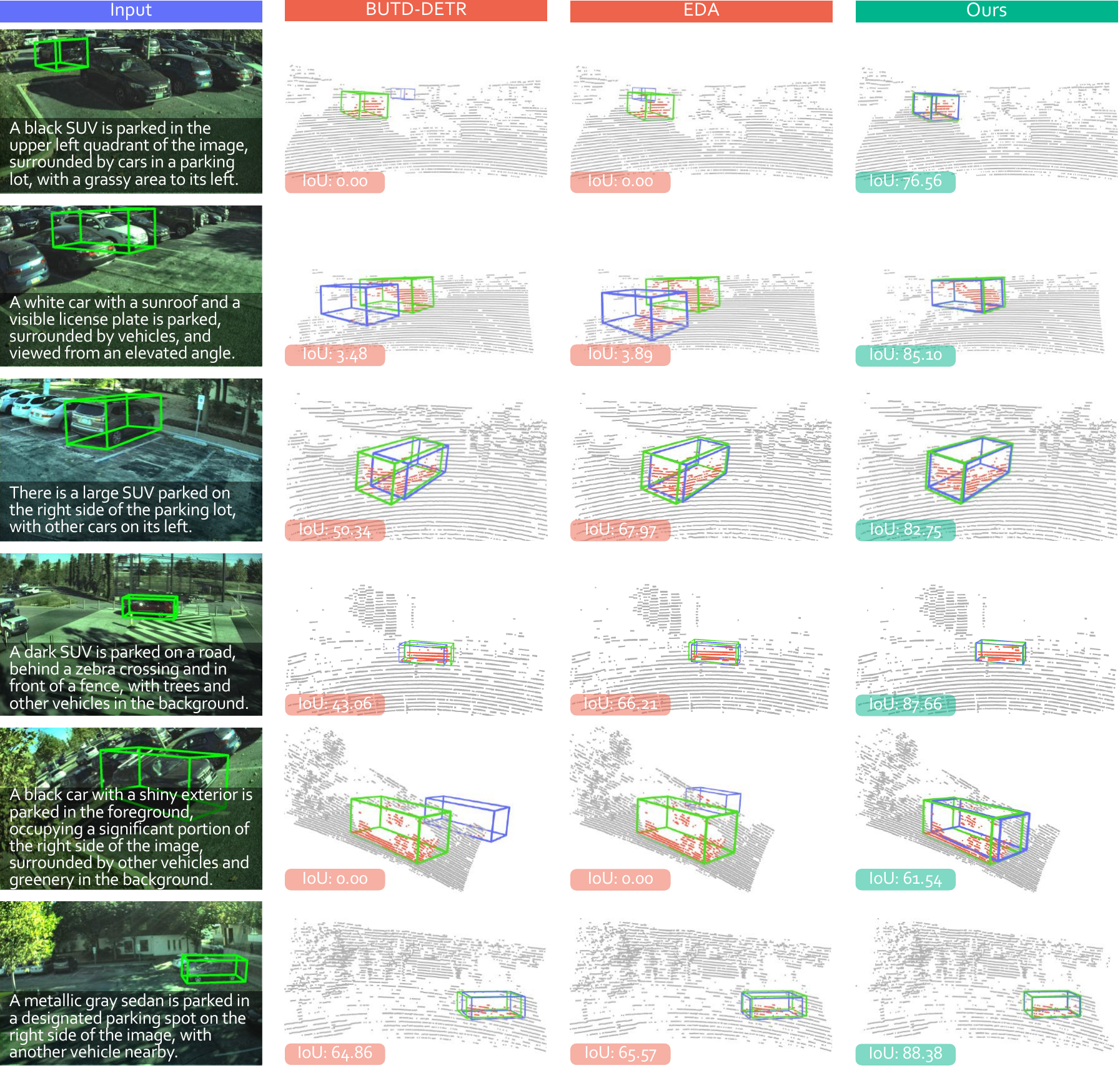}
    \vspace{-0.4cm}
    \caption{\textbf{Additional qualitative comparisons} of single-object 3D grounding on the {\includegraphics[width=0.028\linewidth]{figures/icons/drone.png}} \texttt{Drone} platform from the \ours~dataset. The data shown include the RGB frames, the LiDAR point clouds, and the associated referring expressions. The ground truth and predicted boxes are shown in green and blue, respectively. Best viewed in colors and zoomed in for more details.}
\label{fig:supp_single_drone}
\end{figure}

\begin{figure}[t]
    \centering
    \includegraphics[width=\linewidth]{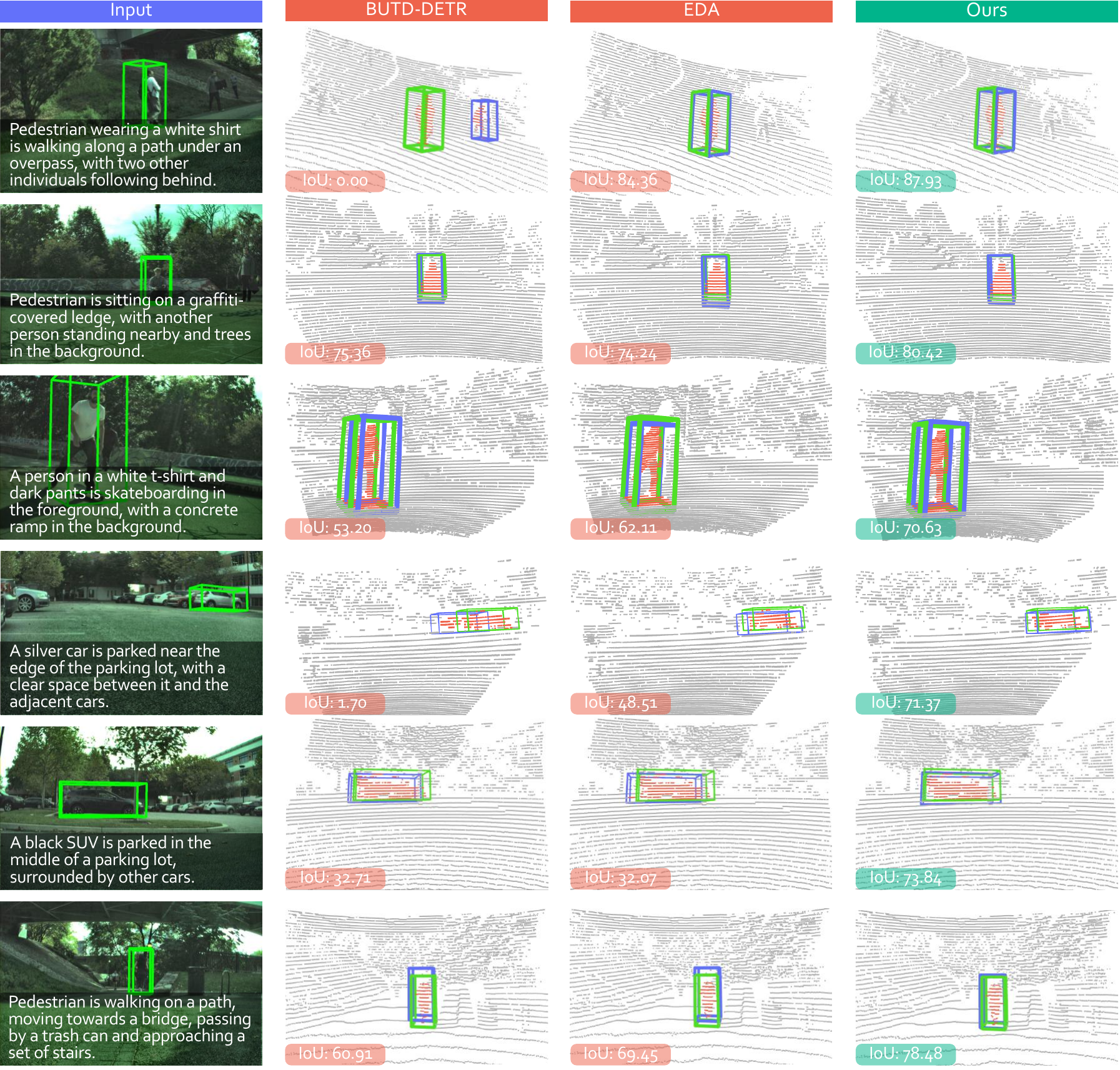}
    \vspace{-0.4cm}
    \caption{\textbf{Additional qualitative comparisons} of single-object 3D grounding on the {\includegraphics[width=0.022\linewidth]{figures/icons/quadruped.png}} \texttt{Quadruped} platform from the \ours~dataset. The data shown include the RGB frames, the LiDAR point clouds, and the associated referring expressions. The ground truth and predicted boxes are shown in green and blue, respectively. Best viewed in colors and zoomed in for more details.}
\label{fig:supp_single_quad}
\end{figure}

\begin{figure}[t]
    \centering
    \includegraphics[width=\linewidth]{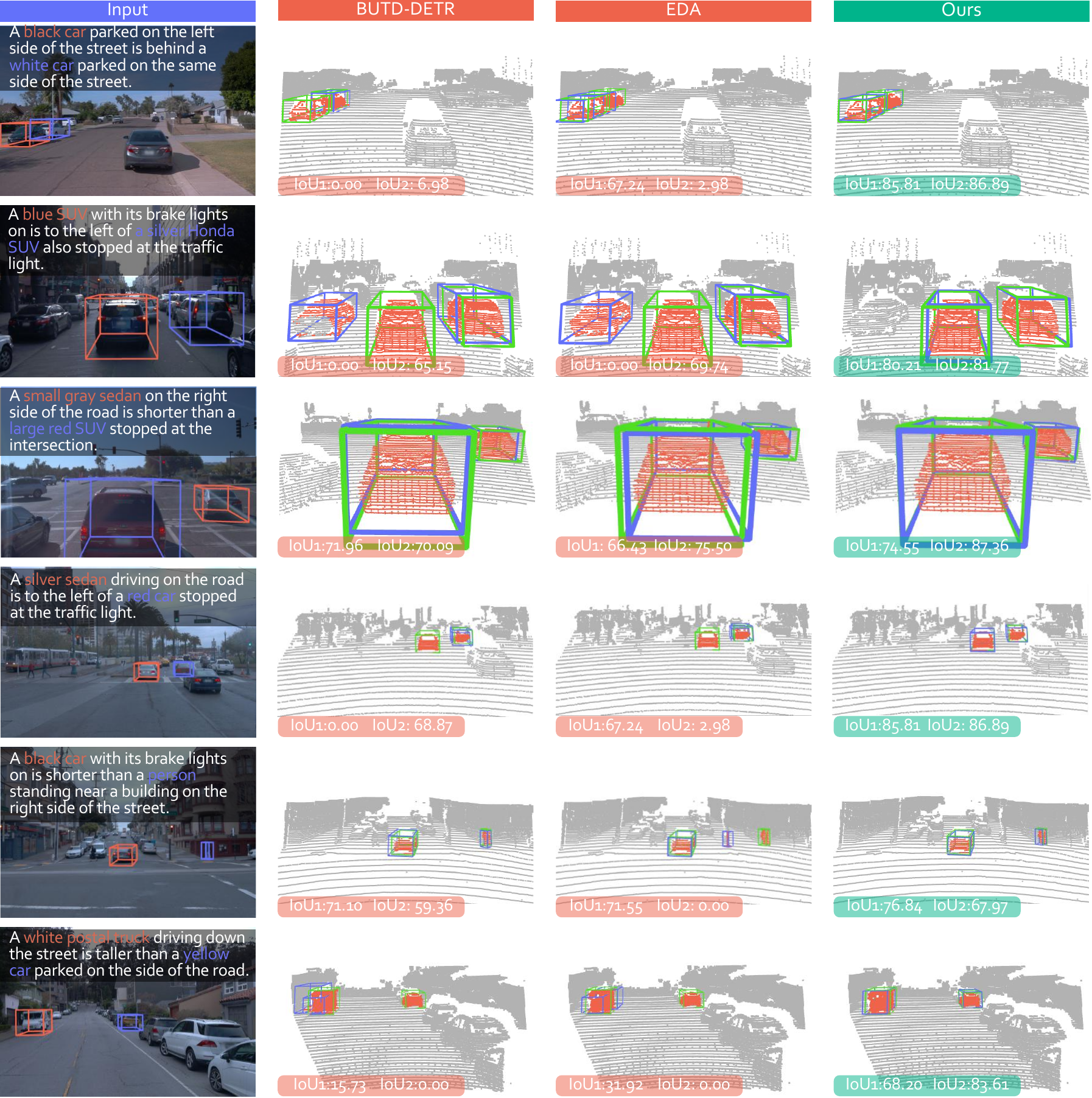}
    \vspace{-0.4cm}
    \caption{\textbf{Additional qualitative comparisons} of multi-object 3D grounding approaches on the \ours~dataset. The data shown include the RGB frames, the LiDAR point clouds, and the associated referring expressions. The ground truth and predicted boxes in the prediction results are shown in green and blue, respectively. Best viewed in colors and zoomed in for more details.}
\label{fig:supp_multi}
\vspace{-0.2cm}
\end{figure}

\section{Additional Visual Comparisons}

In this section, we provide more qualitative examples to complement the main results. These visualizations illustrate the strengths and failure patterns of different methods across sensor platforms and grounding settings.

\subsection{Qualitative Results for Single-Object 3D Grounding}

Figure~\ref{fig:supp_single_vehicle}, Figure~\ref{fig:supp_single_drone}, and Figure~\ref{fig:supp_single_quad} present single-object grounding results from the {\includegraphics[width=0.021\linewidth]{figures/icons/vehicle.png}}~\texttt{Vehicle} {\includegraphics[width=0.028\linewidth]{figures/icons/drone.png}}~\texttt{Drone} and {\includegraphics[width=0.022\linewidth]{figures/icons/quadruped.png}}~\texttt{Quadruped} platforms, respectively. These comparisons reveal several key insights:
\begin{itemize}
    \item \textit{Vehicle Platform} (Figure~\ref{fig:supp_single_vehicle}). Our method consistently localizes referred objects more accurately, particularly in crowded scenes. For instance, in examples involving parked or moving vehicles near intersections, our model correctly resolves spatial descriptions like ``moving forward on the street, positioned near the crosswalk'' or ``parked on the right side of the street'', whereas baseline methods often misplace the box or miss the object entirely.

    \item \textit{Drone Platform} (Figure~\ref{fig:supp_single_drone}). Despite the elevated perspective and sparse point clouds, our method produces robust results by leveraging cross-platform cues. Notably, in scenes with occlusions or dense parking lots, our model successfully grounds phrases like ``black SUV with grassy area to its left'' and ``white car with sunroof'', demonstrating resilience to complex layouts and ambiguous references. In contrast, EDA and BUTD-DETR frequently fail to produce any box or yield inaccurate boundaries.

    \item \textit{Quadruped Platform} (Figure~\ref{fig:supp_single_quad}). Grounding from the quadruped perspective introduces unique challenges due to low-angle views and close-range objects. Our method shows clear improvements, accurately grounding pedestrians and vehicles even when facing away from the camera or interacting with the environment. For example, descriptions such as ``moving towards a bridge'' and ``near the edge of the parking lot'' are correctly localized only by our approach. Baselines either regress coarse boxes or misinterpret perspective cues.
\end{itemize}
These qualitative comparisons validate the platform-agnostic design of our approach and demonstrate the ability to disambiguate fine-grained language in diverse visual-spatial contexts.

\subsection{Qualitative Results for Multi-Object 3D Grounding}

Figure~\ref{fig:supp_multi} illustrates representative examples from the multi-object grounding setting. Here, each scene contains two referred objects and a complex expression that captures both individual characteristics and inter-object relationships.

Our method shows notable advantages in:
\begin{itemize}
    \item \textit{Capturing relative semantics:} In expressions like ``a white oistal truck is taller than a yellow car'' or ``a silver sedan is to the left of a red car'', our model localizes both objects with high precision and correct relative positioning.

    \item \textit{Handling comparatives and prepositions:} Even in cases with overlapping objects or subtle distinctions, our method interprets spatial relations (\eg, ``to the left of'', ``is behind'') more reliably than baselines.

    \item \textit{IoU consistency:} The paired IoU scores (IoU$1$/IoU$2$) of our predictions are consistently higher, reflecting better localization and object differentiation.
\end{itemize}

In contrast, BUTD-DETR \cite{jain2022butd-detr} often fails to detect one of the objects, while EDA \cite{wu2023eda} tends to confuse spatial hierarchy, misplace referred instances, or miss the relationships altogether.

Overall, these visual results demonstrate that our model excels not only in individual object grounding but also in multi-entity reasoning, which is crucial for real-world applications requiring collaborative spatial understanding.

\section{Broader Impact \& Limitations}
In this section, we elaborate on the broader impact, societal influence, and potential limitations.

\subsection{Broader Impact}

This work introduces a new benchmark and methodology for 3D visual grounding across diverse robotic platforms, including vehicles, drones, and quadrupeds. By addressing cross-platform perception and grounding under real-world sparsity, we hope to inspire future research in robust, generalizable spatial language understanding. The dataset and evaluation settings reflect realistic conditions encountered by embodied agents in autonomous driving, inspection, and delivery. We expect this work to benefit the development of safe, context-aware decision-making systems that can interpret human intent across environments. All data collection and annotation followed privacy-compliant and publicly accessible sources.

\subsection{Societal Influence}

The ability to ground language in 3D scenes is critical for real-world human-robot interaction, especially in complex outdoor scenarios. Our benchmark enables evaluating such capabilities beyond indoor or single-device assumptions, pushing toward a more inclusive and scalable understanding. Potential downstream applications include collaborative navigation, voice-based robotics control, and assistive technologies in search-and-rescue operations. While our dataset promotes progress in these areas, we note that grounding models trained on limited sensory conditions may inadvertently inherit biases from pretrained language models or overlook vulnerable populations in data-scarce environments.

\subsection{Potential Limitations}

Despite its scale and diversity, our dataset may still suffer from platform-specific biases (\eg, drone views emphasizing sparse or elevated contexts), which could limit generalization. The current version focuses primarily on static scenes with one or more referred objects, without modeling temporal dynamics or dialogue-based interaction. In addition, our evaluation settings assume accurate text descriptions and do not yet account for ambiguous, contradictory, or noisy language input. Furthermore, while our benchmark covers three robotic platforms, generalization to other types of sensors or modalities (\eg, thermal, event cameras) remains unexplored.

\section{Public Resource Used}
In this section, we acknowledge the use of the public resources, during the course of this work:

\subsection{Public Datasets Used}
\begin{itemize}
    \item M3ED\footnote{\url{https://m3ed.io}.}\dotfill CC BY-SA 4.0

    \item Waymo Open Dataset\footnote{\url{https://github.com/waymo-research/waymo-open-dataset}.} \dotfill Apache License 2.0
\end{itemize}

\subsection{Public Implementation Used}
\begin{itemize}
    \item BUTD-DETR\footnote{\url{https://github.com/nickgkan/butd_detr}.} \dotfill CC BY-SA 4.0 License
    \item EDA\footnote{\url{https://github.com/yanmin-wu/EDA}.} \dotfill CC BY-SA 4.0 License 
    \item Open3D\footnote{\url{http://www.open3d.org}.} \dotfill MIT License  
    \item PyTorch\footnote{\url{https://pytorch.org}.} \dotfill BSD License  
    \item Pointnet2 PyTorch \footnote{\url{https://github.com/erikwijmans/Pointnet2_PyTorch}.} \dotfill UNLICENSE
    \item PointNet++\footnote{\url{https://github.com/charlesq34/pointnet2  }.} \dotfill MIT License
    \item xtreme1\footnote{\url{https://github.com/xtreme1-io/xtreme1}.} \dotfill Apache License 2.0
    \item WildRefer \footnote{\url{https://github.com/4DVLab/WildRefer}.} \dotfill CC BY-SA 4.0 License 
\end{itemize}

\clearpage\clearpage
\bibliographystyle{plain}
{\small\bibliography{main.bib}}

\end{document}